\documentclass[10pt,journal,compsoc]{IEEEtran}

\ifCLASSOPTIONcompsoc
  \usepackage[nocompress]{cite}
\else
  \usepackage{cite}
\fi

\usepackage{lineno}
\usepackage{graphicx} 		
\usepackage{subfigure}	
\usepackage{booktabs}
\usepackage{multirow}
\usepackage{makecell}
\usepackage{enumerate}
\usepackage{ragged2e}
\usepackage{caption}
\usepackage[dvipsnames]{xcolor}
\usepackage[colorlinks,linkcolor=blue]{hyperref}
\usepackage{url}
\usepackage{utfsym}
\usepackage{wasysym}
\usepackage{threeparttable}
\usepackage{color}

\usepackage{tabularx}
\usepackage{enumitem}

\newcommand\sbullet[1][.5]{\mathbin{\vcenter{\hbox{\scalebox{#1}{$\bullet$}}}}}

\usepackage{tikz}
\usetikzlibrary{arrows,shapes,positioning,shadows,trees}
\tikzset{
	basic/.style  = {draw, text width=3.2cm, font=\sffamily, rectangle},
	root/.style   = {basic, rounded corners=2pt, thick, align=center, fill=red!30},
level 2/.style = {basic, rounded corners=6pt, thick, align=center, fill=SeaGreen!60, text width=10em},
	level 3/.style = {basic, thin, align=left, fill=YellowOrange!15, text width=11em}
}

\begin{document}

\title{Time Series Analysis for Education: Methods, Applications, and Future Directions}

\author{
Shengzhong Mao,
Chaoli Zhang$^{\dagger}$,
Yichi Song,
Jindong Wang,
Xiao-Jun Zeng, \\
Zenglin Xu,~\IEEEmembership{Senior~Member,~IEEE},
and Qingsong Wen$^{\dagger}$,~\IEEEmembership{Senior~Member,~IEEE}
	
\IEEEcompsocitemizethanks{\IEEEcompsocthanksitem S. Mao and X. Zeng are with the Department of Computer Science, University of Manchester, Manchester M13 9PL, UK. E-mail: \{shengzhong.mao, x.zeng\}@manchester.ac.uk.
\IEEEcompsocthanksitem C. Zhang is with the School of Computer Science and Technology, Zhejiang Normal University, Zhejiang, China. E-mail: chaolizcl@zjnu.edu.cn.
\IEEEcompsocthanksitem Y. Song is with the Department of Math and Statistics, Carleton College, USA. E-mail: songc2@carleton.edu.
\IEEEcompsocthanksitem J. Wang is with William \& Mary, Williamsburg, VA, USA. E-mail: jwang80@wm.edu.
\IEEEcompsocthanksitem Z. Xu is with Fudan University, Shanghai, China. E-mail: zenglin@gmail.com.
\IEEEcompsocthanksitem Qingsong Wen is with Squirrel Ai Learning, Bellevue, WA, USA. E-mail: qingsongedu@gmail.com.}
\thanks{$^{\dagger}$Corresponding authors: Chaoli Zhang and Qingsong Wen.}
\thanks{\leftline{Project page: \url{https://github.com/ai-for-edu/time-series-analysis-for-education}}}
}

\IEEEtitleabstractindextext{%
\begin{abstract}
\justifying 
Recent advancements in the collection and analysis of sequential educational data have brought time series analysis to a pivotal position in educational research, highlighting its essential role in facilitating data-driven decision-making. However, there is a lack of comprehensive summaries that consolidate these advancements. To the best of our knowledge, this paper is the first to provide a comprehensive review of time series analysis techniques specifically within the educational context. We begin by exploring the landscape of educational data analytics, categorizing various data sources and types relevant to education. We then review four prominent time series methods—forecasting, classification, clustering, and anomaly detection—illustrating their specific application points in educational settings. Subsequently, we present a range of educational scenarios and applications, focusing on how these methods are employed to address diverse educational tasks, which highlights the practical integration of multiple time series methods to solve complex educational problems. Finally, we conclude with a discussion on future directions, including personalized learning analytics, multimodal data fusion, and the role of large language models (LLMs) in educational time series. The contributions of this paper include a detailed taxonomy of educational data, a synthesis of time series techniques with specific educational applications, and a forward-looking perspective on emerging trends and future research opportunities in educational analysis. The related papers and resources are available and regularly updated at: \url{https://github.com/ai-for-edu/time-series-analysis-for-education}.
\end{abstract}

\begin{IEEEkeywords}
Time Series Analysis,
Educational Data Analytics,
Forecasting,
Classification,
Clustering,
Anomaly Detection,
Personalized Learning,
Educational Data Mining
\end{IEEEkeywords}}

\maketitle
\IEEEdisplaynontitleabstractindextext
\IEEEpeerreviewmaketitle

\IEEEraisesectionheading{\section{Introduction}\label{secIntroduction}}
\IEEEPARstart{I}{}n recent years, the educational landscape has witnessed significant transformation, driven by the integration of advanced digital learning technologies and data analytics. 
Given the growing abundance of educational data, characterized by its sequential nature, time series analysis has emerged as a crucial tool in educational research \cite{box2015time,romero2020educational,zhang2024self}, enabling more informed decision-making and personalized learning experiences.
This approach, which examines time-ordered data points, is effective for analyzing a broad spectrum of data sources, from daily attendance records and test scores to more granular data such as clickstream logs from online learning platforms \cite{wang2024research,chango2022review}. These applications allow researchers and educators to uncover underlying patterns, trends, and relationships within educational datasets, providing critical perspectives on various aspects of education.

\begin{figure}[!htbp]
	\centering
	\includegraphics[scale=0.38]{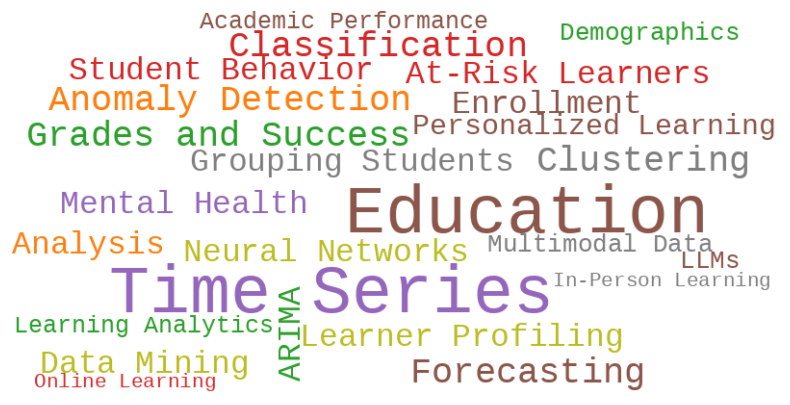}
	\caption{The visualization of key concepts and applications of time series analysis for educational contexts.}
	\label{fig2word}
\end{figure}

Time series methods have become essential for extracting significant observations from inherently temporal educational data, with diverse applications ranging from tracking student performance and engagement to analyzing institutional trends and predicting future outcomes \cite{romero2020educational,batool2023educational,dol2023classification,cerezo2024reviewing,aldowah2019educational}. For instance, by applying time series analysis to historical academic data, educators can anticipate student outcomes and intervene when necessary, thereby improving retention rates and academic success \cite{batool2023educational,roslan2022educational}. Similarly, time series classification of student participation patterns on online learning platforms can help tailor educational content to better meet individual needs, fostering a more personalized learning experience \cite{xu2017progressive,pallathadka2023classification,dol2023classification}. Moreover, by detecting anomalies in time series data related to student behavior, such as sudden drops in grades or engagement, educators can identify at-risk students early, allowing for timely and targeted support \cite{cerezo2024reviewing,bachhal2021educational}. Fig. \ref{fig2word} highlights the key concepts of time series analysis for education.

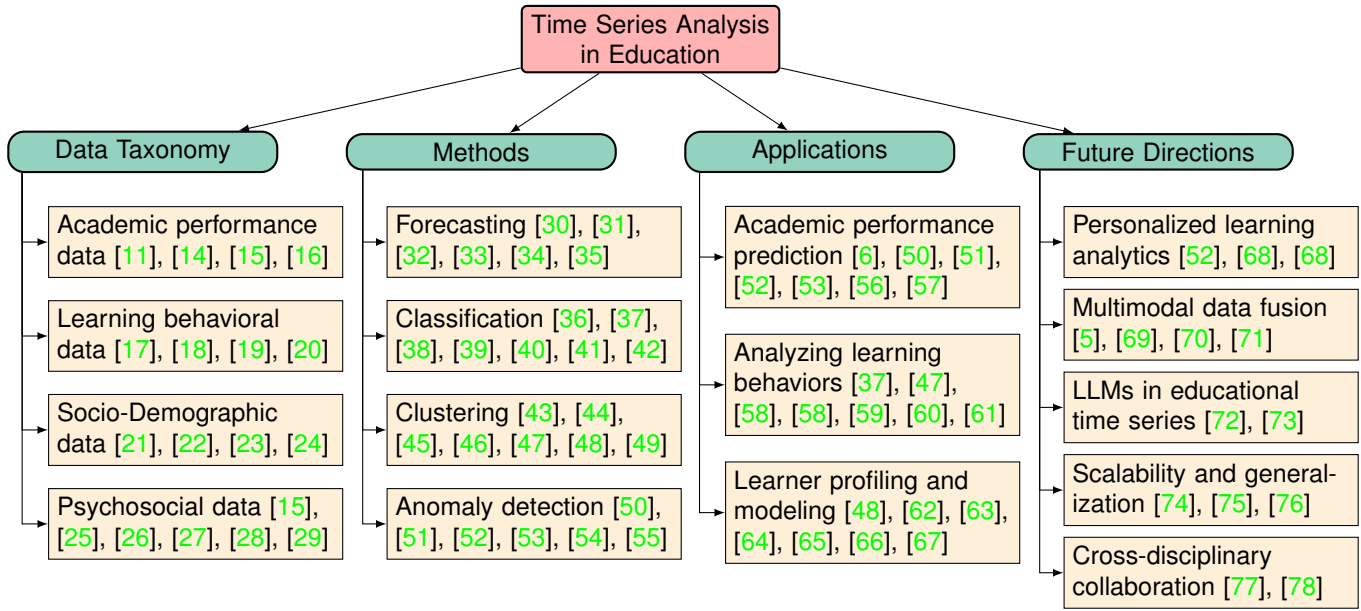
\begin{figure*}[!htbp]
	\centering
	\begin{tikzpicture}[
		level 1/.style={sibling distance=45mm},
		edge from parent/.style={->,draw},
		>=latex]
		
		\node[root] {Time Series Analysis in Education}
		child {node[level 2] (c1) {Data Taxonomy}}
		child {node[level 2] (c2) {Methods}}
		child {node[level 2] (c3) {Applications}}
		child {node[level 2] (c4) {Future Directions}};
		
		\begin{scope}[every node/.style={level 3}]
			\node [below of = c1, xshift=20pt, yshift=-2mm] (c11) {Academic performance data \cite{kamley2016review,xu2017progressive,shingari2017review,helal2019identifying}};
			\node [below of = c11, yshift=-2.5mm] (c12) {Learning behavioral data \cite{gasevic2017detecting,chen2018early,papamitsiou2014temporal,zacharis2015multivariate}};
			\node [below of = c12, yshift=-2.5mm] (c13) {Socio-Demographic data \cite{uddin2017proposing,kiekens2019predicting,costa2017evaluating,shannon2019predicting}};
			\node [below of = c13, yshift=-2.5mm] (c14) {Psychosocial data \cite{shingari2017review,sapiezynski2017academic,ge2020predicting,ruiz2020predicting,yang2017multimodal,ashraf2018comparative}};

			\node [below of = c2, xshift=20pt,yshift=-2mm] (c21) {Forecasting \cite{chen2023comparative,xu2020student,su2018exercise,hussain2019using,beaulac2019predicting,burrus2019predictors}};
			\node [below of = c21, yshift=-2.5mm] (c22) {Classification \cite{paireekreng2015integrated,sheeba2018prediction,valencia2023learning,rao2016predicting,gambo2021artificial,zhang2018students,giannakas2021deep}};
			\node [below of = c22, yshift=-2.5mm] (c23) {Clustering \cite{chakraborty2016density,stites2019cluster,ramos2021approach,wang2017active,akhanli2024hierarchical,cahapin2023clustering,liu2017study}};
			\node [below of = c23, yshift=-2.5mm] (c24) {Anomaly detection \cite{veerasamy2020using,jayaprakash2014early,hong2024early,na2017identifying,guo2022educational,vaidya2024anomaly}};
			
			\node [below of = c3, xshift=20pt,yshift=-4mm] (c31) {Academic performance prediction \cite{veerasamy2020using,batool2023educational,al2019detecting,galici2023close,hong2024early,na2017identifying,jayaprakash2014early}};
			\node [below of = c31, yshift=-7mm] (c32) {Analyzing learning behaviors \cite{			he2023clustering,dutsinma2020vark,he2023clustering,akhanli2024hierarchical,sheeba2018prediction,bueno2023hierarchical,shen2017clustering}};
			\node [below of = c32, yshift=-7mm] (c33) {Learner profiling and modeling \cite{chi2021research,laxhammar2013online,moubayed2020student,xu2019automatic,kuswandi2018k,cahapin2023clustering,trivedi2020clustering}};
			
			\node [below of = c4, xshift=20pt,yshift=-2mm] (c41) {Personalized learning analytics \cite{shoaib2024ai,hong2024early,shoaib2024ai}};
			\node [below of = c41, yshift=-1mm] (c42) {Multimodal data fusion \cite{henderson2020improving,chango2022review,li2020identifying,mu2020multimodal}};
			\node [below of = c42, yshift=-1mm] (c43) {LLMs in educational time series \cite{dan2023educhat,xu2024foundation}};
			\node [below of = c43, yshift=-1mm] (c44) {Scalability and generalization \cite{fu2020learning,govea2023optimization,zhang2024hyperscale}};
			\node [below of = c44, yshift=-1mm] (c45) {Cross-disciplinary collaboration \cite{li2023retracted,xie2024co}};
		\end{scope}
		
		\foreach \value in {1,2,3,4}
		\draw[->] (c1.190) |- (c1\value.west);
		
		\foreach \value in {1,...,4}
		\draw[->] (c2.189) |- (c2\value.west);
		
		\foreach \value in {1,...,3}
		\draw[->] (c3.190) |- (c3\value.west);
		
		\foreach \value in {1,...,5}
		\draw[->] (c4.189) |- (c4\value.west);
	\end{tikzpicture}
         \caption{The hierarchical categorization of the key components of time series analysis in education.}\label{figOverview}
\end{figure*}

Despite the growing importance of time series analysis in education, its applications in current research remain relatively fragmented, with a lack of comprehensive reviews that consolidate the advancements and applications of these methods within educational contexts. Existing research often focuses on specific methodologies or narrow applications—such as forecasting student performance or classifying learning patterns—without offering a comprehensive overview of how time series techniques can be systematically applied to address a broad range of educational challenges \cite{zhang2021educational,yaugci2022educational}. While some review works on educational data mining cover a wider array of educational tasks, there is a notable absence of research that addresses the integration of multiple time series methods to tackle complex sequential educational challenges, as well as the potential of advanced strategies in enhancing time series analysis in educational settings \cite{baek2023educational,aldowah2019educational}.

To address these gaps, as shown in Fig. \ref{figOverview}, this paper provides a systematic overview of time series methods as applied to educational contexts. It aims to synthesize existing methodologies while exploring how these methods can be combined and applied to address multifaceted educational problems. 
Specifically, we begin with an in-depth exploration of educational data sources and types, covering various learning environments and available datasets, and then classify and detail the features of these data related to educational time series, organizing them into four primary categories.
Following this, the paper reviews and discusses key methods of time series analysis covering forecasting, classification, clustering, and anomaly detection, and their specific applications in education. Additionally, it summarizes educational tasks and scenarios from a macro time series perspective, extending beyond specific analytical methods, and outlines future research directions, highlighting the integration of multimodal fusion, large language models (LLMs), and advanced analytical methods.

The main contributions of this paper are as follows:

\begin{itemize}
\item To the best of our knowledge, this paper is the first to comprehensively review time series analysis in educational contexts, covering mainstream methods such as forecasting, classification, clustering, and anomaly detection.
\item The paper emphasizes practical applications in real-world educational scenarios, demonstrating how multiple strategies can be employed simultaneously to improve educational outcomes.
\item This work presents a detailed taxonomy of educational data, categorizing various sources and types, intended to be useful for both current and future applications in educational analysis.
\item We outline future research directions on the integration of advanced techniques in educational time series, aiming to lay the foundation for ongoing and future research in evolving educational analysis.
\end{itemize}

The remainder of this paper is organized as follows:
Section \ref{secRelated} reviews the related works and Section \ref{secData} discusses the types of educational data and analytics.
In Section \ref{secMethod}, we examine the methodologies of time series analysis and their specific applications in educational context. 
Section \ref{secApplication} presents various scenarios and applications in education, and in Section \ref{secFuture}, we explore future directions in the field, highlighting emerging trends and potential advancements in future educational time series analysis. Finally, we provide our conclusions in Section \ref{secConclusion}.

\section{Related Works}\label{secRelated}
Numerous studies have explored the application of time series methods in educational settings, focusing primarily on tasks such as predicting student performance \cite{roslan2022educational,waheed2020predicting,yaugci2022educational,zhang2021educational}, grouping student behaviors \cite{dol2023classification,zhao2023construction,pang2014clustering}, and identifying anomalies in learning processes \cite{vaidya2024anomaly,ren2021deep}. Traditional statistical approaches like the autoregressive integrated moving average (ARIMA) model and exponential smoothing have been widely applied to forecast trends in student grades and attendance \cite{romero2010educational,batool2023educational,sweeney2015next,xu2017progressive,tsiakmaki2018predicting}. More recently, machine learning techniques, including support vector machines (SVM), random forests, and k-nearest neighbors (KNN), artificial neural network (ANN), as well as deep learning models like recurrent neural networks (RNNs) and long short-term memory (LSTM) networks, have demonstrated superior accuracy in various regression and classification tasks \cite{dol2023classification,waheed2020predicting,batool2023educational}. 
Additionally, for student grouping and modeling, techniques such as decision trees and k-means clustering have been employed to identify patterns in student data \cite{chi2021research,wang2024research,saputra2023application,cahapin2023clustering}. For instance, these techniques are used to segment students into distinct categories based on their interaction patterns in virtual learning environments \cite{cerezo2024reviewing,gambo2021artificial,baek2023educational}. Collectively, these studies highlight the impact of time series analysis in addressing diverse educational challenges.

In the broader domain of educational data mining, extensive research has been conducted on leveraging time series data to enhance learning analytics \cite{aldowah2019educational,bachhal2021educational,wang2024research,cerezo2024reviewing}. Subsequent studies have expanded the scope by incorporating broader applications, such as using time series for early warning systems to predict student dropouts or detecting at-risk students based on their interaction patterns in online learning environments \cite{aldowah2019educational,romero2020educational,bachhal2021educational,chango2022review}. While these surveys cover a wide range of educational tasks and scenarios, the integration of multiple time series strategies and the use of advanced techniques such as multimodal fusion and LLMs remain underexplored and warrant further investigation.

Despite the significant progress made by previous literature in applying time series analysis within educational contexts, much of the existing research tends to focus on specific applications or methods. Compared to other influential related surveys from the past five years, as summarised in Table \ref{tabSurvey}, this paper seeks to address these limitations by providing an integrative overview that synthesizes time series methods across a broad range of educational applications. Unlike most surveys, which often concentrate on specific methods or educational tasks and provide only a brief outlook, this paper aims to cover a broader collection of general educational scenarios while offering a more detailed and thorough outlook. Additionally, it integrates the latest techniques, offering potential solutions to current and future multifaceted educational challenges. By doing so, this paper contributes to a deeper understanding of how time series analysis can be leveraged to improve educational outcomes, setting the stage for future research in the dynamic education field.

\begin{table}[!htbp]  
	\centering
	\renewcommand{\arraystretch}{1.4}
	\caption{Comparison of our survey with existing educational surveys (last 5 years). Most surveys typically focus on specific methods or educational tasks and provide a brief overview. Our survey aims to cover a wider array of general educational scenarios and provide a more detailed and comprehensive outlook (F - Forecasting, C - Classification, Cl - Clustering, and AD - Anomaly Detection).}\label{tabSurvey}%
 	   \begin{threeparttable}
	\begin{tabular*}{\hsize}{@{}@{\extracolsep{\fill}}lccccccc@{}} 
		\toprule  
		\multirowcell{2}[-3pt][l]{Survey} & \multicolumn{4}{c}{Methods}& \multicolumn{2}{c}{Applications} & \multirowcell{2}[-3pt][l]{Outlook}\\
		\cmidrule(lr){2-5}  \cmidrule(lr){6-7}       
		& F & C & Cl & AD & Specific & General &  \\ 
		\midrule 
2019 \cite{aldowah2019educational}&\CIRCLE &\LEFTcircle &\LEFTcircle &\Circle &\usym{2714}&&\\   
		2020 \cite{romero2020educational}&\CIRCLE &\LEFTcircle &\Circle &\LEFTcircle &&\usym{2714}&\\   
		2021 \cite{bachhal2021educational}&\CIRCLE &\LEFTcircle &\LEFTcircle &\LEFTcircle &&\usym{2714}&\\   
		2021 \cite{zhang2021educational}&\CIRCLE &\CIRCLE &\CIRCLE &\Circle &\usym{2714}&&\usym{2714}\\  
  		2022 \cite{chango2022review}&\CIRCLE &\LEFTcircle &\Circle &\Circle &&\usym{2714}&\usym{2714}\\     
		2022 \cite{roslan2022educational}&\CIRCLE &\LEFTcircle &\LEFTcircle &\Circle &\usym{2714}&&\\     
2022 \cite{yaugci2022educational}&\CIRCLE &\CIRCLE &\Circle &\Circle &\usym{2714}&&\\    2023 \cite{baek2023educational}&\CIRCLE &\Circle &\LEFTcircle &\Circle &\usym{2714}&&\\  
2023 \cite{batool2023educational}&\CIRCLE &\CIRCLE &\LEFTcircle &\Circle &\usym{2714}&&\\   
2023 \cite{dol2023classification}&\LEFTcircle &\CIRCLE &\CIRCLE &\Circle &\usym{2714}&&\usym{2714}\\   
2024 \cite{cerezo2024reviewing}&\CIRCLE&\Circle&\Circle&\Circle&
        \usym{2714}& &\usym{2714} \\     2024 \cite{wang2024research}&\Circle&\CIRCLE&\Circle&\CIRCLE&&
        \usym{2714}& \\
		\midrule
		Ours    &\CIRCLE       &\CIRCLE       &\CIRCLE       &\CIRCLE       &       &\usym{2714}        &\usym{2714}   \\ 
		\bottomrule    
	\end{tabular*}%
    \begin{tablenotes}
   	\item $\Circle$ represents “not covered”, $\LEFTcircle$ represents “partially covered”, and $\CIRCLE$ represents “fully covered”.
\end{tablenotes}
\end{threeparttable}
\end{table}

\section{Educational Data and Analytics}\label{secData}
Educational data is crucial for analyzing and improving the learning process. With the advent of advanced data collection technologies and the widespread adoption of digital learning environments, the volume and diversity of educational data have increased exponentially. These data span various educational settings, including traditional in-person classrooms, online learning platforms, and blended learning environments \cite{romero2013data,romero2014survey}.
Educational data analytics involves the systematic collection, analysis, and interpretation of data from these diverse educational contexts. It reveals critical information about student performance, learning behaviors, and overall educational outcomes. This section explores the domain of educational data, offering a structured overview that lays the foundation for applying time series analysis in this field. It is designed to provide a comprehensive understanding of the sources and taxonomy of educational data, which is essential for the subsequent analysis and applications discussed in this paper.

\begin{table*}[!htbp]
	\tiny
	\centering
	\caption{Educational data collected from three main learning environments: in-person learning, online learning, and blended learning. The data includes various formats, including time series structures and multimedia types (video, audio, text). Categories of data and their capture methods are detailed according to the specified classification framework.}\label{tab1multidata}
	\renewcommand{\arraystretch}{1.2}
	\resizebox{\textwidth}{!}{
		\begin{tabular}{l|l|lll|l}
			\hline
			\multicolumn{1}{l}{Environment} & \multicolumn{1}{l}{Data Format} & \multicolumn{1}{l}{Name} & \multicolumn{1}{l}{Category} & \multicolumn{1}{l}{Capture} &  \multicolumn{1}{l}{References}\\
			\hline			
			\multirowcell{12}[0pt][l]{In-Person\\ learning}&\multirow{9}{*}{Time series} & Student heart rate & Physical & Sensor & 
\multirowcell{9}[0pt][l]
{				\cite{romero2013data,mu2020multimodal,giannakos2019multimodal}\\
\cite{daoudi2021improving,gadaley2020classroom,henderson20194d}\\
\cite{olsen2020temporal,monkaresi2016automated,worsley2014multimodal}
} \\
			&& Blood volume & Physical & Sensor &  \\
			&& Electrodermal activity & Physiological & Sensor &  \\
			&& Body temperature & Physiological & Sensor &  \\
			&& Hands movement & Physical & Sensor &  \\
			&& Student posture & Physical & Sensor &  \\
			&& Following movement & Physical & Sensor &  \\
			&& Student head attention & Physical & Webcam &  \\
			&& Gaze-based cognitive load & Physical & Sensor &  \\
			\cline{2-6} 			
			&\multirowcell{3}[0pt][l]{Video/Audio\\/Text} & Student behavior  & Physical & Webcam&
\multirowcell{3}[0pt][l]
{\cite{mao2019classroom,gadaley2020classroom,olsen2020temporal}\\
\cite{prieto2018multimodal,henderson20194d,ma2015research}
} \\
			&&Teacher presentation & Physical & Microphone&\\
			&& Writing activity & Digital & Platform &\\
			\hline
			\multirowcell{11}[0pt][l]{Online\\ learning}&\multirow{8}{*}{Time series} & Moodle course activities & Digital & Platform &
\multirowcell{8}[0pt][l]
{\cite{henderson20194d,henderson2020improving,liao2019exploring}\\
\cite{di2017learning,tian2018learning,sharma2019stimuli}\\
\cite{hussain2011affect,liu2019learning}
} \\
			&& Dynamic mouse records & Digital & Log& \\
			&& Evaluation records & Digital & CSV &\\
			&& Student scores & Digital & Log &\\
			&& Eye movement & Physical & Sensor &\\
			&& Electrocardiogram & Physiological & Sensor& \\
			&& Facial electromyogram & Physiological & Sensor& \\
			&& Weather condition & Environmental & Platform &\\
			\cline{2-6}
			&\multirowcell{3}[0pt][l]{Video/Audio\\/Text} & Teacher behavior  & Physical & Webcam&
\multirowcell{3}[0pt][l]
{\cite{wu2020recognition,peng2021recognition,luo2022three}\\
\cite{liu2019learning,tian2018learning,hussain2011affect}
} \\
			&& Student thoughts & Digital & Platform &\\
			& & Speech between students & Physical & Microphone\\
			\hline
			\multirowcell{10}[0pt][l]{Blended\\ learning} & \multirow{7}{*}{Time series} & Teacher body movement & Physical & Sensor & 
\multirowcell{7}[0pt][l]
{\cite{chen2016hybrid,bahreini2016data,li2020incremental}\\
\cite{qu2021research,shankar2019architecture,chango2021multi}\\
} \\
			&       & Teacher joint positions & Physical & Sensor &  \\
			&       & Student online interactions & Digital & Platform &  \\
			&       & Online teaching data & Digital & Log   &  \\
			&       & Offline teaching data & Digital & Log   &  \\
			&       & Digital tool adaptors & Digital & CSV   &  \\
			&       & IoT adaptors & Digital & CSV   &  \\
			\cline{2-6}
			&\multirowcell{3}[0pt][l]{Video/Audio\\/Text}& Facial emotion detection & Physical & Webcam & 
\multirowcell{3}[0pt][l]
{
\cite{chango2021multi,bahreini2016data,qu2021research}\\
\cite{xu2019automatic,chango2021improving}\\
} \\
			& & Teacher speech & Digital & Log   &  \\
			& & Student evaluation & Digital & Log   &  \\
			\hline 
	\end{tabular}}
\end{table*}

\subsection{Educational Data Sources}
Educational data is gathered from a range of learning environments, each providing distinct perspectives on the educational process. Understanding these environments and the types of data they produce is significant for effective educational data analytics. 
In this subsection, we introduce three primary of learning environments. Besides, various educational data formats collected from these settings are discussed, including time series data and multimedia types (video/audio/text), based on their structural characteristics and unique features.
This categorization forms the basis for analyzing temporal patterns in educational domains, which is one primary focus of this paper. Moreover, it presents opportunities for integrating multimodal data, such as combining time series data with video and text, indicating a promising direction for the future of multimodal educational data analytics \cite{giannakos2019multimodal,chango2022review}.

\subsubsection{Learning Environments}
This part introduces three categories of learning environments for collecting educational data: in-person learning, online learning, and blended learning. 
Specifically, as illustrated in Table \ref{tab1multidata}, each data source generated from these environments is comprehensively detailed, including data format, category, capture method (e.g., camera, microphone, logs, etc.), and so forth. Here the categories include five types according to the educational data classification framework \cite{mu2020multimodal,chango2022review}: digital, physical, physiological, psychological, and environmental measurements.

\vspace{0.5em}
\noindent \textbf{In-Person Learning.} 
It refers to the traditional classroom learning that occurs within a physical classroom setting where students and instructors interact face-to-face. This method of learning fosters immediate feedback, spontaneous discussions, and hands-on activities, which helps develop communication skills, collaboration, and a sense of community among students.
Data collection in this environment includes attendance records, student attributes, behavioral observations, and physical interactions \cite{romero2013data}. Additionally, more sophisticated methods such as classroom sensors and wearable devices can provide data on student engagement and interaction patterns. The data generated in in-person learning environments is often qualitative, consisting of teacher notes and observational records, as well as quantitative, including test scores and attendance figures \cite{mu2020multimodal}.

As described in Table \ref{tab1multidata}, the time series features of student physical and physiological data are a major focus in the study of in-person teaching and learning. 
Researchers have extensively examined variables such as heart rate, body temperature, and electrodermal activity to understand student physiological responses and stress levels during classes \cite{giannakos2019multimodal,daoudi2021improving,gadaley2020classroom}. Additionally, in-class student behaviors, particularly movements and attention levels as monitored by various sensors, are widely analyzed to gauge engagement and participation \cite{henderson2020improving}. Several studies have integrated time series data with video and text, examining elements such as student posture, writing activities, and facial expressions \cite{andrade2016using}. Others have employed a broad spectrum of data sources, including audio recordings, images, and video captured by cameras, to create a multi-faceted view of the learning environment \cite{olsen2020temporal,monkaresi2016automated}. This allows for a richer and more detailed analysis of student engagement, attentiveness, and participation over time, helping to identify patterns and correlations that can inform teaching strategies.
While the majority of research has concentrated on student-collected data, some studies have also considered data from instructors \cite{prieto2018multimodal}. For instance, teacher presentations recorded via microphones, along with data from wearable devices by educators, are analyzed to gain insights into teaching practices and their effectiveness \cite{gadaley2020classroom,worsley2014multimodal}. The incorporation of time series data from instructors includes variables like speech patterns, movement around the classroom, gestural communication, and interaction frequencies with students. 

\vspace{0.5em}
\noindent \textbf{Online Learning.}
Online learning, also known as e-learning or virtual learning, takes place through digital platforms where students and instructors interact remotely. Unlike traditional classroom settings, online education leverages digital technologies to facilitate teaching and learning processes, often through platforms that support video lectures, interactive modules, and various forms of communication \cite{luo2022three,wu2020recognition}. This mode of learning not only allows for asynchronous learning but also enables real-time interaction between instructors and students across different geographical locations. The diverse methodologies and data collection techniques employed in online learning environments bring to light essential aspects of student engagement, behavior, and performance, contributing to a more personalized and effective educational experience.

In online education, various time series features are employed to capture dynamic and evolving student interactions and physiological responses. 
A significant source of this data is the digital records from online course activities, which chronicle student engagement with course materials over time, providing a detailed log of participation \cite{sharma2019stimuli}. 
Similarly, dynamic mouse records, captured through digital logs, track real-time mouse movements and clicks, revealing patterns of interaction and levels of user engagement \cite{peng2021recognition}.
In addition, evaluation records and digital logs of student scores offer a chronological account of assessments and academic achievements \cite{liu2019learning,liao2019exploring}. Eye movement data, which tracks where and how long students focus on different parts of the screen, combines with electrocardiogram data collected by physiological sensors to provide indicators of stress and emotional responses during learning activities \cite{henderson20194d}. Environmental data, such as weather conditions, is also considered, as it can influence student mood and cognitive function, thereby affecting engagement with studies \cite{di2017learning}.
Beyond time series features, online learning environments incorporate various other data types. For example, the analysis of speech between students facilitates the understanding of peer interactions and collaborative learning, contributing to a comprehensive view of student thoughts and the overall educational experience.

\vspace{0.5em}
\noindent \textbf{Blended Learning.}
This combines traditional face-to-face classroom methods with online educational practices, aiming to leverage the best of both worlds. This approach integrates the direct interaction and hands-on experiences of in-person learning with the flexibility and technological advantages of online education \cite{qu2021research}. For example, a student might attend a lecture in person and then complete assignments online, allowing them to engage with the material at their own pace and revisit concepts as needed \cite{chango2021multi}.
In a blended learning environment, students benefit from the immediate feedback and social interaction of a physical classroom, which fosters a sense of community and collaboration. This setting allows for real-time discussions, group work, and direct support from instructors, enhancing the overall learning experience. 

Time series data in blended learning encompass both physical and digital sources that provide insights into both teacher and student behaviors. These data are employed to analyze the dynamic interplay between in-person and online educational activities. 
Physical time series features include teacher body movement and joint positions \cite{xu2019automatic,wu2020recognition}. These measurements help optimize teaching methods by analyzing how instructors physically interact with students and the classroom environment. Additionally, facial emotion detection provides real-time insights into student emotional states, which can inform adjustments to teaching strategies to enhance engagement and learning outcomes \cite{bahreini2016data}.
On the digital side, time series features such as student online interactions, online teaching data, and offline teaching data are critical. These datasets, logged through various digital platforms, record how students engage with both online and offline components of the course \cite{qu2021research}. By examining patterns in online activities, such as forum participation, assignment submissions, and content access, educators learn student engagement and learning behaviors. Besides, teacher speech and student evaluations further contribute to understanding the effectiveness of instructional methods and student feedback \cite{wu2020recognition,chango2021improving,li2020incremental}.
Recent studies incorporate data from digital tool adaptors and IoT adaptors, facilitating the integration of digital tools and IoT devices \cite{shankar2019architecture,chen2016hybrid}. For instance, data from IoT devices can provide real-time feedback on environmental conditions, such as classroom temperature and lighting, which can affect student comfort and concentration. 
While in-person learning predominantly focuses on physical and physiological data, and online learning emphasizes digital interaction data, blended learning integrates both. This hybrid approach allows for a more broad analysis of the educational experience, capturing the nuances of how students transition between and interact with both physical and virtual learning environments.

\begin{table*}[!htbp]
	\tiny
	\centering
	\caption{Four main categories of educational data repositories. For each category, the table provides details on the datasets, including data format, sample size, publication year, available links, and the primary educational tasks they support (F - Forecasting, C - Classification, Cl - Clustering, AD - Anomaly Detection, TS - Time Series).}\label{tab2datasource}
	\renewcommand{\arraystretch}{1.4}
	\resizebox{\textwidth}{!}{
		\begin{tabular}{llllll}
			\hline
			Data Source & Dataset  & Tasks & Data Format & Sample/Feature & Year/Link \\
			\hline
			\multirowcell{5}[0pt][l]{UCI ML\\Repository}
			& Student Performance Dataset & F, C & TS & 649/30 & 2008/\href{https://archive.ics.uci.edu/dataset/320/student+performance}{Link} \\
			& User Knowledge Modeling Dataset & C, Cl & TS & 403/5 & 2013/\href{https://archive.ics.uci.edu/dataset/257/user+knowledge+modeling}{Link} \\
			& Educational Process Mining Dataset & F, C, Cl & TS/Text & 230,318/13 & 2015/\href{https://archive.ics.uci.edu/dataset/346/educational+process+mining+epm+a+learning+analytics+data+set}{Link} \\
			& OULAD & F, C, Cl & TS/Text & 32,593/43 & 2017/\href{https://archive.ics.uci.edu/dataset/346/educational+process+mining+epm+a+learning+analytics+data+set}{Link} \\
			& Student Academics Performance & C & TS & 300/22 & 2018/\href{https://archive.ics.uci.edu/dataset/467/student+academics+performance}{Link} \\
			\hline
			\multirowcell{6}[0pt][l]{Mendeley\\Repository} 
			& KEEL Dataset Repository & F, C, Cl & TS/Text & 39,602/40 & 2018/\href{https://data.mendeley.com/datasets/py4hhv3rb8/1}{Link} \\
			& MOOC Lectures Dataset & F, C, Cl & TS/Video/Text & 12,032/40 & 2019/\href{https://data.mendeley.com/datasets/xknjp8pxbj/1}{Link} \\
			& Flip Teaching in Physics Lab Data & F, C, Cl & TS & 1,233/4 & 2019/\href{https://data.mendeley.com/datasets/68mt8gms4j/3}{Link} \\
			& OBE Dataset & F, C & TS/Text & 34,650,000/20 & 2019/\href{https://data.mendeley.com/datasets/9zkfwdm8xf/1}{Link} \\
			& Influencing Factors of Teacher Burnout & F, C & TS & 876/5 & 2020/\href{https://data.mendeley.com/datasets/6jmv43nffk/2}{Link} \\
			& Academic Performance Evolution Data & F & TS & 12,411/44 & 2020/\href{https://data.mendeley.com/datasets/83tcx8psxv/1}{Link} \\
			\hline
			\multirowcell{6}[0pt][l]{Harvard\\Dataverse} 
			& HarvardX Person-Course & F, C & TS/Text & 338,223/20 & 2013/\href{https://doi.org/10.7910/DVN/26147}{Link}
 \\
			& MOOC-Ed Network Dataset & F, C & TS/Text & 6,052/13 & 2015/\href{https://doi.org/10.7910/DVN/ZZH3UB}{Link}
 \\
			& CAMEO Dataset & C, AD & TS/Text &  1,893,092/12 & 2015/\href{https://doi.org/10.7910/DVN/3UKVOR}{Link}
 \\
			& Canvas Network Open Courses & Cl & TS/Text & 325,000/25 & 2016/\href{https://doi.org/10.7910/DVN/1XORAL}{Link}
 \\
			& Video Game Learning Analytics & F, C & TS/Video/Audio & 331/25 & 2020/\href{https://doi.org/10.7910/DVN/V7E9XD}{Link}
 \\
			& Interdisciplinary Student Dataset & C & TS/Text & 807/29 & 2020/\href{https://doi.org/10.7910/DVN/M07HQ7}{Link}
 \\
			\hline
			\multirowcell{6}[0pt][l]{Educational\\Competitions} 
			& KDD Cup 2010 & F, C & TS/Text & 9,353/20 & 2010/\href{https://pslcdatashop.web.cmu.edu/KDDCup/}{Link}
 \\
			& KDD Cup 2015 & F, C & TS & 120,543/22 & 2015/\href{http://moocdata.cn/challenges/kdd-cup-2015}{Link} \\
			& NAEP 2017 ASSISTments Competition & F, C & TS/Text & 942,817/76 & 2017/\href{https://sites.google.com/view/assistmentsdatamining/home}{Link} \\
			& NAEP 2019 Educational Competition & F, C & TS/Text & 438,392/7 & 2019/\href{https://sites.google.com/view/dataminingcompetition2019/home}{Link} \\
			& EdNet Dataset & F, C & TS & 131,417,236/12 & 2019/\href{https://github.com/riiid/ednet}{Link} \\
			& Riiid AIEd Challenge 2020 & C & TS & 99,271,300/22 & 2020/\href{https://www.kaggle.com/c/riiid-test-answer-prediction/}{Link} \\
			\hline
			\multirowcell{5}[0pt][l]{Miscellaneous\\Sources} 
			& DataSchop@CMU & F, C, Cl, AD & TS/Text & 8,526,184/36 & 2010/\href{https://pslcdatashop.web.cmu.edu/}{Link}\\
			& NUS Multisensor Presentation & F, C & TS/Video/Audio & 36,000/51 & 2015/\href{https://scholarbank.nus.edu.sg/handle/10635/137261}{Link}\\
			& Learn Moodle August 2016 & F, C & TS/Text & 6,119/20 & 2016/\href{http://research.moodle.net/158/}{Link}\\
			& MUTLA Dataset & F, C, Cl & TS/Video/Text & 114,977/40 & 2018/\href{https://github.com/RyanH98/SAILData}{Link}\\
			& Junyi Academy Dataset & F, C, Cl & TS/Video/Text & 16,217,311/36 & 2019/\href{https://www.kaggle.com/datasets/junyiacademy/learning-activity-public-dataset-by-junyi-academy}{Link}\\
			\hline
	\end{tabular}}%
\end{table*}%

\subsubsection{Publicly Available Datasets}
This section provides an overview of widely accessible educational datasets used in the research, categorized by their sources. These include the UCI Machine Learning Repository, Mendeley Data Repository, Harvard Dataverse, Educational Competitions, and other miscellaneous sources. 
Table \ref{tab2datasource} provides detailed information about each dataset, including its source, applied tasks, data formats, and additional relevant details.

\vspace{0.5em}
\noindent {\textbf{UCI ML Repository.}} 
The UCI ML Repository is a key source for machine learning datasets, including many from the educational sector. These datasets are primarily suitable for the forecasting, classification, and clustering tasks.

\begin{itemize}
\item \textbf{(i)} \href{https://archive.ics.uci.edu/dataset/320/student+performance}{Student Performance Dataset} \cite{cortez2008using}. It contains 649 samples and 30 features, including demographic, social, and school-related factors, focusing on estimating student end-of-term scores.
\item \textbf{(ii)} \href{https://archive.ics.uci.edu/dataset/257/user+knowledge+modeling}{User Knowledge Modeling Dataset} \cite{kahraman2013development}. It includes 403 samples with five features centered on learning activities, and it is suitable for classification and clustering, enabling predictions of learners knowledge levels.
\item \textbf{(iii)} \href{https://archive.ics.uci.edu/dataset/346/educational+process+mining+epm+a+learning+analytics+data+set}{Educational Process Mining Dataset} \cite{vahdat2015learning}. It comprises 230,318 instances described by 13 features from the activities of 115 students using an educational simulator, which supports applications like forecasting, classification, and clustering.
\item \textbf{(iv)} \href{https://archive.ics.uci.edu/dataset/349/open+university+learning+analytics+dataset}{Open University Learning Analytics Dataset (OULAD)} \cite{kuzilek2017open}. 
It includes course, student, and interaction data for seven modules and is used to identify at-risk students, predict engagement, and analyze demographics in online education.
\item \textbf{(v)} \href{https://archive.ics.uci.edu/dataset/467/student+academics+performance}{Student Academics Performance} \cite{hussain2018educational}. It consists of 300 records described by 24 features, focusing on demographic factors. It is mainly used for classification tasks to predict student performance.
\end{itemize}
	
\vspace{0.5em}
\noindent {\textbf{Mendeley Data Repository.}}
This cloud-based platform is essential for managing academic data, including time series, video, and text features. It hosts a diverse collection of educational data that supports various studies and analyses.

\begin{itemize}
\item \textbf{(i)} \href{https://data.mendeley.com/datasets/py4hhv3rb8/1}{Knowledge Extraction Based on Evolutionary Learning (KEEL) Data} \cite{hou2019anti}. It is part of the KEEL project, and this repository includes many datasets compatible with KEEL software, facilitating a broad range of knowledge discovery tasks.
\item \textbf{(ii)} \href{https://data.mendeley.com/datasets/xknjp8pxbj/1}{MOOC Lectures Dataset} \cite{kastrati2020wet}. It includes word embeddings and topic vectors from transcripts of 12,032 video lectures across 200 Coursera courses.
\item \textbf{(iii)} \href{https://data.mendeley.com/datasets/68mt8gms4j/3}{Flip Teaching in Physics Lab Data} \cite{gomez2020effectiveness}. It examines performance of 1,233 engineering students in Physics and Electricity courses from 2013 to 2017, comparing traditional and flip teaching methods.
\item \textbf{(iv)} \href{https://data.mendeley.com/datasets/9zkfwdm8xf/1}{Outcome-Based Education (OBE) Dataset} \cite{joseresearch}. It includes 34,650,000 entries across 20 programs, structured into 21 files, which assesses outcome-based education in an engineering college.
\item \textbf{(v)} \href{https://data.mendeley.com/datasets/6jmv43nffk/2}{Influencing Factors of Teacher Burnout} \cite{prasojo2020teachers}. It includes responses from 876 teachers in a survey, investigating correlations between self-concept and efficacy as predictors of burnout.
\item \textbf{(vi)} \href{https://data.mendeley.com/datasets/83tcx8psxv/1}{Academic Performance Evolution Data} \cite{delahoz2020dataset}. It describes performance evolution for 12,411 engineering students, including academic, social, and economic data across 44 features.
\end{itemize}

\vspace{0.5em}
\noindent {\textbf{Harvard Dataverse Repository.}}
The Harvard Dataverse houses 98,873 datasets across 13 disciplines. It offers educational data in formats like time series, video, and text, used for tasks like forecasting and anomaly detection. Its user-friendly features are invaluable for researchers.

\begin{itemize}
\item \textbf{(i)} \href{https://doi.org/10.7910/DVN/26147}{HarvardX Person-Course} \cite{ho2015harvardx,DVN/26147_2014}. It contains 338,223 entries with 20 features, useful for analyzing user progress, access patterns, and predicting course outcomes.
\item \textbf{(ii)} \href{https://doi.org/10.7910/DVN/ZZH3UB}{MOOC-Ed Network Dataset} \cite{kellogg2015massively}. It is designed to examine dropout rates in MOOCs and explore self-regulated learning behaviors. 
\item \textbf{(iii)} \href{https://doi.org/10.7910/DVN/3UKVOR}{CAMEO Dataset} \cite{northcutt2016detecting}. It contains information on student activities in MITx and HarvardX courses, used to study copying answers through multiple online accounts. 
\item \textbf{(iv)} \href{https://doi.org/10.7910/DVN/1XORAL}{Canvas Network Open Courses} \cite{DVN/1XORAL_2016}. It has been available since 2016, and this collection comprises 325,000 records with 25 features. Each entry details activities of individuals in 238 courses, commonly used for clustering tasks.
\item \textbf{(v)} \href{https://doi.org/10.7910/DVN/V7E9XD}{Video Game Learning Analytics} \cite{DVN/V7E9XD_2020}. It includes an early reading and writing assessment in preschool settings, using video game learning analytics. This Comprises data from 331 students, highlighting phonological awareness, literacy activities, and test scores. 
\item \textbf{(vi)} \href{https://doi.org/10.7910/DVN/M07HQ7}{Interdisciplinary Student Dataset} \cite{DVN/M07HQ7_2020} It is for the situated academic writing self-efficacy scale validation project, including 807 observations with 29 variables. It covers demographic information from 543 undergraduate and 246 graduate students.
\end{itemize}

\vspace{0.5em}
\noindent \textbf{Educational Competitions.} 
This part focuses on publicly accessible datasets from educational competitions originating from prestigious conferences and various hackathons. These datasets offer immediate validation for proposed models and competitive rankings against other participants.

\begin{itemize}
	\item \textbf{(i)} \href{https://pslcdatashop.web.cmu.edu/KDDCup/}{The KDD Cup} \cite{stamper2010bridge}. Launched in 2010, this competition tasks participants with predicting student performance using logs of interactions with tutoring systems. The 2015 edition focused on forecasting dropouts in XuetangX MOOCs, China's largest MOOC platform.
	\item \textbf{(ii)} \href{https://sites.google.com/view/assistmentsdatamining/home}{The NAEP Competition} \cite{naep2017,naep2019}. The 2017 competition used deidentified click-stream data from middle school students using ASSISTments, while the 2019 edition provided datasets to predict student activities based on early test data.
	\item \textbf{(iii)} \href{https://github.com/riiid/ednet}{EdNet Dataset} \cite{choi2020ednet}. Collected over two years by Santa, an AI-driven tutoring service in Korea with 780,000 users, this hierarchical dataset includes four sub-datasets focused on logged actions, ideal for deep learning applications.
	\item \textbf{(iv)} \href{https://www.kaggle.com/c/riiid-test-answer-prediction/}{Riiid AIEd Challenge 2020} \cite{tran2021riiid}. Hosted on Kaggle, this challenge features a dataset of 418 lectures and 170 questions, with actions from 393,656 users. Participants develop knowledge tracing models to predict future student performance.
\end{itemize}

\vspace{0.5em}
\noindent \textbf{Miscellaneous Sources.} 
In addition to the widely recognized repositories and competitions, there are several miscellaneous sources that provide valuable educational datasets. These sources encompass a variety of platforms and organizations that offer datasets for diverse educational research needs.

\begin{itemize}
	\item \textbf{(i)} \href{https://pslcdatashop.web.cmu.edu/}{DataShop@CMU} \cite{koedinger2010data}. A comprehensive repository for learning science researchers, offering secure data storage and various analytical tools. It hosts 40 datasets with detailed, longitudinal data across multiple semesters and courses.
	\item \textbf{(ii)} \href{https://scholarbank.nus.edu.sg/handle/10635/137261}{NUS Multisensor Presentation} \cite{gan2015multi}. It contains time series, video, and audio data from 51 individuals recorded at the National University of Singapore, and it is used to provide feedback on oral presentation skills and identify presentation patterns.
	\item \textbf{(iii)} \href{http://research.moodle.net/158/}{Learn Moodle August 2016} \cite{moodle2016}. An anonymized dataset from the Learn Moodle MOOC, capturing user activities. It includes records of badges issued, course completions, grades, and event logs.
	\item \textbf{(iv)} \href{https://github.com/RyanH98/SAILData}{MUTLA Dataset} \cite{xu2019mutla} A multimodal dataset for teaching and learning analytics, including user records, brainwave data, and webcam footage, aimed at analyzing teacher-student interactions.
	\item \textbf{(v)} \href{https://www.kaggle.com/datasets/junyiacademy/learning-activity-public-dataset-by-junyi-academy}{Junyi Academy Dataset} \cite{chang2015modeling}. It contains student interaction data with the Junyi Academy platform, detailing log-in times, study durations, exercise types, and response accuracy.
\end{itemize}
				
\begin{table}[!htbp]
	\centering
	\caption{Educational data taxonomy including four main categories. For each category, specific data types are identified along with examples to illustrate the nature of the data.}\label{tab3taxomony}
	\renewcommand{\arraystretch}{1.15}
	\begin{tabular*}{\hsize}{@{}@{\extracolsep{\fill}}lll@{}}
		\toprule
		Taxonomy & Data Type & Examples \\ 
		\midrule
		\multirowcell{4}[0pt][l]{Academic\\Performance} 
		& Academic metrics&  $\sbullet[.8]$ exams, quizzes, midterms \\ 
		& Test evaluation &  $\sbullet[.8]$ tests, subject assessments \\ 
		& Assignment marks & $\sbullet[.8]$ homework, lab reports \\ 
		& Attendance records & $\sbullet[.8]$ presence in classes, labs \\  
		\midrule
		\multirowcell{4}[0pt][l]{Learning\\Behavioral}
		& Engage metrics & $\sbullet[.8]$ learning time, frequency \\ 
		& Interaction data & $\sbullet[.8]$ clickstream, interactions \\ 
		& Study patterns & $\sbullet[.8]$ study logs, resource use \\ 
		& Collaboration data & $\sbullet[.8]$ group work, talks \\ 
		\midrule
		\multirowcell{4}[0pt][l]{Socio-\\Demographic}
		& Demographic info & $\sbullet[.8]$ age, gender, nationality \\ 
		& Economic status &  $\sbullet[.8]$ income, education \\ 
		& Family background & $\sbullet[.8]$ structure, involvement \\ 
		& Geographical data & $\sbullet[.8]$ location, school district \\ 
		\midrule
		\multirowcell{4}[0pt][l]{Psychosocial}
		& Mental health & $\sbullet[.8]$ stress, anxiety, depression \\ 
		& Social interactions & $\sbullet[.8]$ peer, teacher relations \\ 
		& Emotional state & $\sbullet[.8]$ emotional monitoring \\ 
		& Motivation metrics & $\sbullet[.8]$ goal-setting, perseverance \\ 
		\bottomrule
	\end{tabular*}
\end{table}

\subsection{Educational Data Taxonomy}\label{secDataTaxonomy}
Educational data can be categorized into various types based on the nature of the information and its application in time series analysis. This subsection introduces four main categories \cite{bai2021educational,guo2022educational}: academic performance data, learning behavioral data, socio-demographic data, and psychosocial data, summarized in Table \ref{tab3taxomony}. This taxonomy provides a structured framework that categorizes data based on characteristics and relevance to educational outcomes. Each category covers specific types of data, offering a comprehensive taxonomy used throughout the paper to align with various time series analysis applications.

\vspace{0.5em}
\noindent\textbf{Academic Performance Data.} 
This refers to all metrics directly related to student achievement and outcomes \cite{kamley2016review,xu2017progressive}. This includes grades from exams, quizzes, and assignments, test scores from standardized tests and subject-specific assessments, and assignment marks from homework, projects, and lab reports \cite{shingari2017review,tamhane2014predicting,sweeney2015next,meier2015predicting}. Attendance records, which log student presence in classes, labs, and other educational activities, also fall under this category \cite{kaviyarasi2018exploring,helal2019identifying,tsiakmaki2018predicting}. Academic performance data are crucial for predicting future student performance, identifying those who may be struggling, and tailoring educational interventions to support student success \cite{hassan2019students}.

\vspace{0.5em}
\noindent\textbf{Learning Behavioral Data.} 
It captures the actions and interactions of students within various learning environments \cite{chen2018early,yang2017behavior,brinton2016mining}. Engagement metrics, such as the time spent on learning platforms, frequency of logins, and participation in discussions, provide valuable insights into student engagement \cite{xu2017progressive,carter2017blending,dvorak2016online}. Interaction data from online learning platforms, including clickstream data and interactions with educational resources, as well as study patterns, such as logs of study sessions and resource usage patterns, are critical components of this category \cite{gasevic2017detecting,hart2017individual,papamitsiou2014temporal,zacharis2015multivariate,ashraf2018comparative}. Additionally, collaboration data, which records group work, peer-to-peer interactions, and other collaborative activities, helps in understanding how students work together \cite{adejo2018predicting,waheed2020predicting}. Learning behavioral data are essential for analyzing student engagement, identifying effective learning strategies, and personalizing learning experiences to meet individual student needs \cite{mubarak2021visual,sandoval2018centralized}.

\vspace{0.5em}
\noindent\textbf{Socio-Demographic Data.} 
It relates to the background and demographic characteristics of students \cite{stewart2019predicting,yang2019demographical}. This includes demographic information such as age, gender, ethnicity, and nationality, as well as socioeconomic status indicators like parental income, education level, and employment status \cite{natek2014student,pandey2016towards,xu2017predicting}. Family background data, including family structure and parental involvement, and geographical data, such as location and school district information, are also part of this category \cite{uddin2017proposing,kiekens2019predicting,costa2017evaluating,shannon2019predicting}. Socio-demographic data is used to analyze the impact of these factors on educational outcomes, develop targeted support programs, and address educational inequalities, providing a comprehensive understanding of the broader context in which students are learning.

\vspace{0.5em}
\noindent\textbf{Psychosocial Data.} 
This involves the psychological and social aspects that influence student behavior and performance. This includes mental health surveys that capture responses about stress, anxiety, depression, and overall mental well-being \cite{sapiezynski2017academic,yang2017multimodal}. Data on social interactions, such as peer relationships and student-teacher interactions, as well as emotional state tracking through applications that monitor emotional states, are also included \cite{shingari2017review,ashraf2018comparative,ge2020predicting}. Additionally, motivational metrics, such as self-reported data on motivation levels, goal-setting, and perseverance, provide insight into the internal factors driving student performance \cite{baker2010better,ruiz2020predicting}. Psychosocial data are crucial for identifying students in need of mental health support, understanding the impact of social factors on learning, and creating interventions to improve overall student well-being.

\begin{figure}[!htbp]
	\centering
	\includegraphics[scale=0.46]{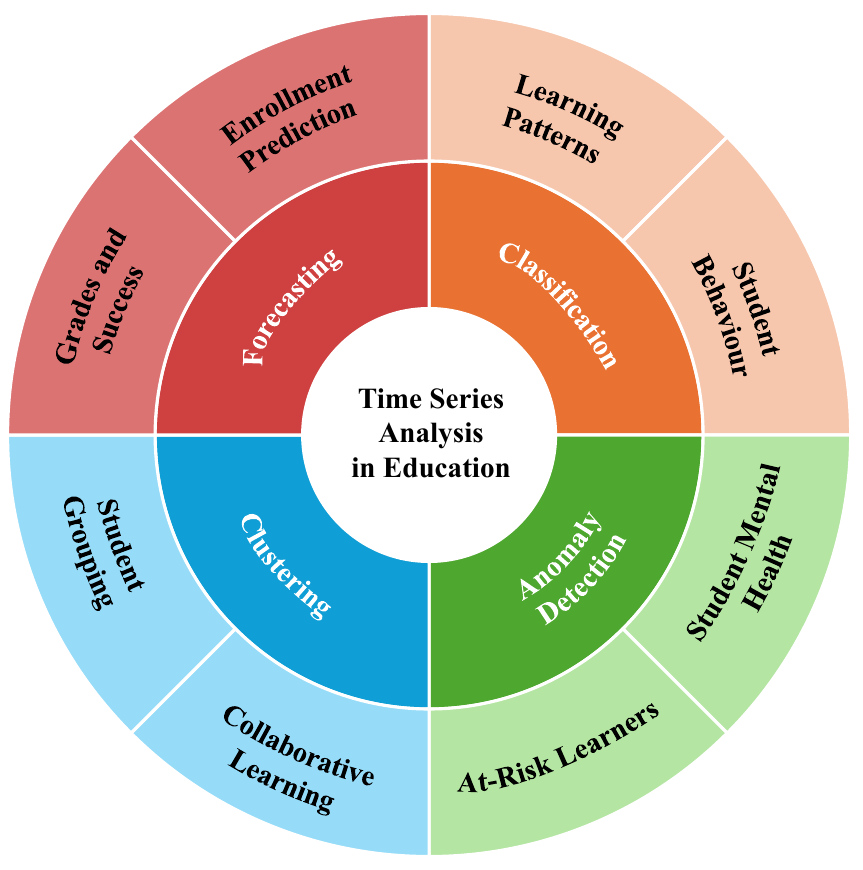}
	\caption{The overview of time series analysis in educational contexts. It highlights four fundamental methods: forecasting, classification, clustering, and anomaly detection, each with its specific applications in educational settings.}
	\label{figMethods}
\end{figure}

\section{Time Series Analysis in Education}\label{secMethod}
Time series analysis is increasingly recognized for its capacity to transform educational research and practice by providing a deep look into temporal data patterns. In education, time series methods help in understanding and predicting trends, behaviors, and outcomes over time, thereby enabling educators, administrators, and policymakers to make data-driven decisions \cite{rezaee2018application,haiyang2018time,dobashi2022learning}. This section introduces and elaborates on four primary time series methods—forecasting, classification, clustering, and anomaly detection—highlighting their specific applications in educational contexts, as presented in Fig. \ref{figMethods}. These methods not only enhance the analytical capabilities within educational data but also drive significant improvements in teaching, learning, and administration. By examining how these methods are utilized, we can discover valuable applications and guide future research in this evolving field.

\subsection{Time Series Forecasting in Education}
Time series forecasting in education predicts future trends and patterns in educational metrics based on historical data.
Educational forecasting spans a range of applications, from predicting individual student performance to estimating institutional enrollment.
Accurate forecasting supports students with personalized assistance and early interventions, while helping educators and administrators optimize resources and make informed decisions \cite{asadi2023ripple}.
Given the distinctive predictive objectives, this section explores two primary aspects of forecasting in education: (i) forecasting grades and success, and (ii) forecasting enrollment.

\subsubsection{Forecasting Grades and Success}
Student performance prediction, especially in terms of grades and academic success, stands as a fundamental task within the educational context. 
By analyzing past performance data, such as test scores, behavioral patterns, and other relevant indicators, educators can gain valuable insights into student learning trajectories and potential academic outcomes \cite{albreiki2021systematic}.
Specifically, forecasting grades involves predicting individual academic performance by estimating future scores in specific subjects or courses, whereas forecasting success focuses on program-level achievement by categorizing students into broader performance levels \cite{kim2018gritnet,xu2020student}.
In this part, we examine the existing body of literature regarding the application, highlighting their significance in enhancing academic performance and supporting student success.

Forecasting grades typically involves predicting or estimating student grades at the course level based on features of academic performance data and learning behavioral data, such as attendance records, assignments, and test scores \cite{su2018exercise,chen2023comparative}. Educators use this process to anticipate student upcoming performance in assessments or courses, aiming to predict the numerical or letter grades they are likely to achieve.
Generally, predictive models for forecasting grades incorporate internal assessments, such as quiz scores, homework, midterms, and other internal evaluations, which serve as significant predictors of final grades \cite{yaugci2022educational}. For instance, researchers have developed predictive models by incorporating attendance records, homework grades, and quiz scores as key variables to forecast student final outcomes for a first-year engineering course at a university \cite{marbouti2015building,cui2019predictive,marbouti2016models}. The continuous nature of these performance indicators makes them critical components in constructing accurate predictive models, providing ongoing feedback throughout the course duration \cite{liu2019ekt,bilal2022role,chen2023comparative}.
Additionally, data from learning management systems, which record detailed student activity logs including login frequencies, assignment submissions, resource accesses, and interactions in discussion forums, are instrumental \cite{tjandra2022student,yang2018study}. These behavioral data points are robust indicators of student engagement and effort, which have been shown to correlate positively with academic performance.

Like program-level prediction, forecasting success involves predicting broader academic outcomes such as overall GPA, graduation rates, and program completion \cite{beaulac2019predicting,burrus2019predictors,hussain2019using}. It often revolves around classifying student performance into discrete categories such as pass-fail, below or above a certain threshold, or broader classifications like good, average, or poor performanc, and often relies on academic background features \cite{alyahyan2020predicting,xu2020student}.
Previous achievements, high school type, admission test scores, and the type of access (regular or special admission programs) are critical predictors of long-term academic success \cite{uddin2017proposing,zeineddine2021enhancing}.
In addition to traditional performance and attendance variables, socio-demographic and psychosocial data have been incorporated to enhance predictive models \cite{shoaib2024ai}. Specifically, socio-demographic factors such as age at admission, gender, socio-economic status, city of origin, and ethnicity are commonly used to establish a baseline understanding of student background \cite{alsariera2022assessment,qiu2022predicting}. Psychological factors, including personality traits like openness and conscientiousness, alongside social networking data such as the number of Facebook friends and types of posts, provide deeper insights into student personality and social behavior, which can influence their academic journey \cite{pallathadka2023classification,wu2022sgkt}.
In some studies, linguistic features extracted from admission essays or other written materials, such as topic modeling, word count, and part-of-speech distribution, have also been utilized to predict student retention and success \cite{ogihara2017student,asselman2023enhancing}.

\subsubsection{Forecasting Enrollment}
Enrollment forecasting is the practice of predicting the number of students expected to enroll in particular courses, programs, or educational institutions in upcoming terms. This forecasting is vital for effective institutional planning and resource management, allowing educational institutions to optimize class sizes, allocate faculty appropriately, and efficiently use facilities \cite{aksenova2006enrollment}. 
Several factors significantly influence student enrollment patterns and, consequently, the accuracy of enrollment forecasts. To forecast enrollment accurately, institutions must draw on a diverse array of data sources, primarily including academic performance and socio-demographic data \cite{cirelli2018predictive,yang2021predicting}. This covers historical enrollment records and previous academic metrics such as high school grades and standardized test scores \cite{trytten2017moving}. Additionally, demographic trends, such as population growth and migration patterns, significantly influence enrollment numbers \cite{whiton2018high,xie2023novel}. Socio-economic indicators like family income and parental education levels are also crucial \cite{aiken2020predicting}. For example, studies have shown that students with higher academic performance in high school are less likely to require remedial courses in college \cite{nakhkob2016predicted,hanif2023study}. Moreover, institutional factors such as the institution's reputation, the availability of desired programs, and the quality of facilities and faculty play a significant role \cite{shao2022machine,slim2018predicting}. By integrating these diverse data points, predictive models can capture the complex factors that impact student enrollment, resulting in more precise and actionable forecasts.

The application of time series forecasting in enrollment prediction has become increasingly prominent, effectively enhancing the accuracy of predicting student enrollment trends through various methodologies \cite{ujkani2021machine}. Educational institutions utilize these models to forecast enrollment for each school and grade, aiding in resource planning and minimizing disruptions at the start of the school year \cite{furstenberg2007encouraging,tanner2021predicting}. Traditional time series techniques like ARIMA and exponential smoothing have proven effective, with these works showcasing the ability of ARIMA model to capture trends and seasonal variations accurately \cite{qin2019impact,calvo2020higher,slim2018predicting}. The integration of machine learning algorithms, such as SVM and ANN, further refines predictive accuracy by modeling non-linear relationships in enrollment data \cite{al2024university,soltys2021enrollment,shao2022machine}. In higher education, these predictive models are pivotal in identifying students at risk of dropping out, enabling early interventions and improving retention rates. Hybrid models, combining multiple forecasting techniques like ARIMA and neural networks, enhance prediction performance by leveraging both linear and non-linear data patterns \cite{gerasimovic2018enrollment}. Moreover, predictive analytics in forecasting admission test outcomes influence enrollment predictions, illustrating how these integrated forecasts enhance overall accuracy and strategic decision-making \cite{slim2018predicting}. Overall, accurate enrollment forecasting plays a crucial role in student success and institutional efficiency, underscoring its broad impact on educational settings.

\subsection{Time Series Classification in Education}
Time series classification is an effective approach used to label sequences of data points over time, thereby identifying patterns that can inform predictions and decision-making processes. In the educational domain, time series classification involves analyzing longitudinal data to uncover trends, behaviors, and events that occur over time within educational contexts \cite{govstautaite2022multi}. This technique is instrumental in understanding student learning and engagement by identifying and interpreting patterns and trends that emerge over time. This part explores the application of classifying learning patterns and student behavior, highlighting the techniques used and their practical implications.

\subsubsection{Classifying Learning Patterns}
Classifying learning patterns include analyzing sequences of educational data to identify different types of learning behaviors and trajectories. This can help in personalizing education, identifying students who may need additional support, and optimizing learning strategies. Educational data typically used in this application includes academic performance data, learning behavioral data, and psychosocial data \cite{xu2017progressive,khamparia2020association,chang2009learning}. 
Firstly, academic metrics like grades and test scores are crucial indicators of student learning patterns. Analyzing temporal data from academic records enables educators to identify trends and patterns in student performance over time. For example, it helps reveal whether a student is struggling with the material or facing external challenges that impact their learning progress \cite{zhang2020learning,brinton2016mining}. 
In addition, behavioral factors, such as the frequency and duration of logins, participation in online discussions, and completion rates of assignments and quizzes, indicate how students interact with educational materials \cite{govstautaite2022multi}. By classifying these behavioral patterns, educators can identify students who are highly engaged and those who may be at risk of falling behind. A student who frequently logs in and participates actively in discussions is likely following a positive learning trajectory, whereas irregular login patterns and incomplete assignments might signal disengagement \cite{rasheed2021learning,qiu2016modeling}.
Psychosocial data includes information related to students motivation, and attitudes towards learning. Survey responses about students stress levels or self-reported levels of motivation can be analyzed over time to identify patterns that may affect academic performance and engagement \cite{dutsinma2020vark,paireekreng2015integrated}.

To classify these learning patterns, various methods are employed. Supervised learning techniques such as decision trees (DT), random forests (RF), and SVM are commonly used \cite{ozpolat2009automatic,azzi2020robust}. These methods rely on labeled training data to learn the mapping between input student features and output types of learning patterns. For instance, decision trees can help in breaking down complex time series data into simpler, interpretable rules that classify different learning patterns \cite{bernard2017learning,ocepek2013exploring}. Random forests, being an ensemble method, improve the robustness and accuracy of the classification by combining multiple decision trees\cite{hasibuan2016detecting,khan2021student}.
Deep learning methods, especially RNNs and LSTMs, are particularly effective to capture temporal dependencies and long-range patterns in sequential data. They learn the temporal dynamics of student behavior and performance over time, and are often used in adaptive learning systems to dynamically adjust the content and difficulty based on the classified learning patterns \cite{valencia2023learning,gambo2021artificial,giannakas2021deep}.
Existing works have demonstrated the efficacy of these methods in various educational contexts. For example, in MOOCs, studies have used SVMs and LSTMs to classify engagement patterns, predicting course completion and dropout rates, which allows for the development of interventions to increase student retention \cite{el2019fuzzy,pallathadka2023classification}. In K-12 education, research has utilized decision trees and random forests to classify reading and math skills development patterns in elementary students, leading to more effective individualized education plans \cite{anitha2015proposing,sheeba2018prediction}.

\subsubsection{Classifying Student Behavior}
Classifying student behavior entails examining various activities, including attendance, participation, and engagement with learning materials \cite{adejo2018predicting}. This process helps to understand and categorize students who exhibit behaviors that correlate with academic success or failure. The educational data typically used in this application includes learning behavioral data and socio-demographic data \cite{jaboob2024integration,qureshi2023factors}. In this task, learning behavioral data includes detailed logs of student activities such as attendance records, frequency of participation in class discussions, time spent on different learning activities, and patterns of accessing and using educational resources \cite{chen2018early}. Analyzing these behaviors over time can help classify students into different behavioral categories, such as highly engaged, moderately engaged, and disengaged students \cite{waheed2020predicting,hwang2012pilot}. Besides, by incorporating the socio-demographic information about student backgrounds, such as age, socioeconomic status, and other demographic factors into classification models, educators can better understand how these factors influence student behavior and engagement patterns \cite{mubarak2021visual}.

Applications of classifying student behavior include the implementation of behavioral interventions, where schools can address negative behaviors identified through classification \cite{ashraf2018comparative,qiu2016modeling}. For instance, a pattern of frequent absences can trigger an intervention to understand and address the underlying causes \cite{yathongchai2013learner}. Engagement analysis is another application, allowing educators to identify when students are most and least engaged, and adjust teaching methods accordingly to maintain or increase engagement \cite{sandoval2018centralized}. Additionally, classification algorithms can detect changes in behavior that may indicate mental health issues, prompting early support from counselors or psychologists.
In higher education, the classification techniques have been used to identify students at risk of dropping out based on their engagement patterns with online learning platforms \cite{pandey2024unobtrusive,tran2020toward}. In blended learning environments, research has shown that classifying patterns of interaction can help predict academic success and guide the development of more effective teaching strategies \cite{deeva2022predicting,zhao2023construction,yang2017behavior}. Techniques used for these classifications include supervised learning methods like decision trees and support vector machines, as well as deep learning methods, which are particularly effective in capturing temporal dependencies in educational time series data \cite{rao2016predicting,zhang2018students,chiu2021bayesian,dutt2019can,dumdumaya2019exploring}. By classifying learning patterns and student behavior, educators can gain valuable insights into how students learn and behave, leading to more effective and personalized educational experiences. 

\subsection{Time Series Clustering in Education}
Time series clustering in education is concerned with analyzing longitudinal data related to student performance, interactions and engagement to form meaningful groups. This process groups time-dependent data into clusters, where data points within each cluster are similar to each other and distinct from those in other clusters \cite{wang2019clustering}. The primary aim of clustering in education is to enhance the learning experience by identifying trends and patterns that can inform teaching strategies and improve educational outcomes. 
By capturing the dynamic nature of student learning features over time, it provides a comprehensive view of the educational process and fosters collaboration by identifying opportunities to group students with complementary learning trajectories and behaviors.

\subsubsection{Clustering Students}
Clustering students aims to categorize students based on various profile information properties, facilitating tailored educational strategies to meet diverse learning needs. 
This typically uses a combination of academic performance data, such as grades, test scores, and assignment completion rates, and learning behavioral data, such as attendance, participation in class activities, and engagement with online learning platforms \cite{kuswandi2018k,chang2020analysis}. Techniques like k-means, hierarchical clustering, and density-based spatial clustering of applications with noise (DBSCAN) are commonly employed for this purpose \cite{lee2016hierarchical,trivedi2020clustering}. Specifically, k-means clustering is a popular algorithm that partitions students into $k$ distinct clusters based on similarities in their data, such as grades, participation levels, or learning styles. This technique is straightforward yet powerful, enabling educators to quickly identify and categorize students who share common characteristics \cite{sari2017implementation,bueno2023hierarchical}. Hierarchical clustering, on the other hand, builds a hierarchy of clusters, which is particularly useful for understanding the relationships between different student groups. This method can reveal subgroups within a larger student population, providing deeper insights into student diversity \cite{oviedo2016hierarchical,akhanli2024hierarchical}. DBSCAN is another technique that is effective in handling noise or outliers in the data, identifying clusters of students with similar behaviors or outcomes, even when the data is messy or contains anomalies \cite{chakraborty2016density,wang2017active,cahapin2023clustering}.

The practical applications of student clustering are broad and impactful. One of the primary uses is in personalized learning, where students are clustered based on their learning styles or performance data \cite{azarnoush2013toward,klingler2016temporally}. This clustering enables educators to design instruction that is tailored to the specific needs of each group. For instance, students who struggle with a particular topic can be grouped and provided with additional resources or alternative teaching methods, thereby improving their chances of success \cite{kinnebrew2013contextualized,moubayed2020student}. 
Another significant application of clustering in education is in optimizing group work. By clustering students with complementary strengths and weaknesses, educators can form balanced and effective teams for collaborative projects \cite{pasina2019clustering}. This ensures that each group has a mix of abilities, which can lead to more productive and enriching group work experiences. For example, a study that clustered students based on their engagement metrics—such as time spent on assignments and participation in class discussions—demonstrated that providing customized support to these groups led to improved academic outcomes \cite{aggarwal2019application,iam2017generating,akbar2018improving}. Similarly, clustering students according to their learning styles allows instructors to adapt their teaching methods to suit these preferences, resulting in enhanced learning experiences and better educational outcomes \cite{nalli2021comparative,pang2014clustering}.

\subsubsection{Clustering Collaborative Learning}
Clustering collaborative learning focuses on grouping students based on their interactions and collaboration patterns in group activities, projects, and discussions. This approach is essential in both traditional and online learning environments, as it fosters teamwork and allows for diverse perspectives to be shared among group members.
Educational data used in this task typically includes information from learning behavioral and psychosocial variables \cite{stites2019cluster,liu2017study,ramos2021approach,cahapin2023clustering}. Learning behavioral data covers the frequency and quality of interactions among group members, participation in group tasks, and contributions to discussions. The metrics such as the number of messages exchanged, response times, and the balance of contributions among group members provide insights into collaborative dynamics \cite{pang2014clustering,perera2008clustering,chakraborty2016density}. Additionally, psychosocial data, gathered through surveys and assessments, capture student attitudes towards group work, teamwork skills, and interpersonal dynamics \cite{wang2017active}. 

Traditional face-to-face learning environments frequently organized into groups with either similar or diverse knowledge levels to effectively capture a wide range of semantic insights.
For instance, previous works introduced the techniques to  form heterogeneous categorises automatically by analyzing the differences among students \cite{bueno2023hierarchical,trivedi2020clustering}.
Collaborative learning is evidently valuable even in online educational settings, where organizing students according to shared interests and preferences boosts both engagement and overall satisfaction. For instance, the k-means along with the weighting strategy are utilized to create more cohesive and interactive learning experiences \cite{sari2017implementation,li2021unsupervised}. Beisdes, spectral clustering is particularly useful for identifying clusters in complex collaborative networks, such as those found in large-scale online learning environments. In these environments, students often interact with multiple peers across various activities, creating a dense and interconnected network of interactions \cite{akhanli2024hierarchical,bueno2023hierarchical,vsaric2023student,bogarin2014clustering,ding2017student}. This method is especially valuable in online courses where face-to-face interaction is limited, and it helps educators recognize well-functioning groups and those that may require reorganization or additional support \cite{oviedo2016hierarchical}.
Overall, clustering techniques has wide applications in collaborative learning such as improving group dynamics and identifying collaboration patterns \cite{saputra2023application,lee2016hierarchical}. By identifying and analyzing interaction patterns within student groups, it not only improve the overall effectiveness of collaborative learning but also contribute to a more engaging and productive learning experience for all students \cite{de2014monitoring,bahel2021student,fan2016clustering,sari2017implementation}.

\subsection{Time Series Anomaly Detection in Education}
Anomaly detection in educational data involves identifying unusual patterns or outliers that deviate from the norm \cite{laxhammar2013online,guo2022educational}. This technique is crucial for recognizing potential issues early, allowing for timely interventions to support students. Anomalies in educational contexts can signify various challenges, from academic struggles to behavioral issues and mental health concerns \cite{vaidya2024anomaly}. By leveraging time series analysis, educators can detect these anomalies, providing a foundation for proactive measures to enhance student outcomes. This section reviews the methods and applications of anomaly detection in education, with a focus on identifying at-risk learners and monitoring student mental health.

\subsubsection{Detecting At-Risk Students}
Detecting at-risk learners involves utilizing various types of educational data to identify students who may face academic difficulties or are at risk of dropping out \cite{jayaprakash2014early,hong2024early}. The primary data categories include academic performance data, learning behavioral data, and socio-demographic data, each providing unique insights.
Academic performance data, such as grades and test scores, directly reflects student achievements. Temporal patterns in these data can highlight significant changes, indicating potential struggles. The performance decrease often predicts future academic failure, allowing educators to intervene before performance deteriorates further \cite{guo2022educational,vaidya2024anomaly}.
Behavioral factors like attendance records and engagement metrics are also critical. Regular attendance and active participation are strong indicators of student success. Combining attendance data with academic performance metrics has proven effective in predicting dropout rates \cite{na2017identifying,veerasamy2020using,galici2023close}. Students with inconsistent attendance are at higher risk, prompting timely interventions.
Socio-demographic data, including socio-economic status, family education levels, and ethnicity, adds context to academic and behavioral data. Although not as directly indicative of academic risk, this data helps understand broader factors influencing student educational experience \cite{hu2020towards}. Integrating socio-demographic data with academic and behavioral data allows for more targeted and effective interventions.

In practical applications, these data types are integrated and analyzed using advanced time series methods to develop early warning systems, continuously monitoring student data and flagging at-risk individuals based on predefined patterns \cite{baneres2020early,al2019detecting,atif2020perceived,wahdan2022early}. Particularly, machine learning strategies are broadly used to analyze combined data to identify unusual patterns indicating risk \cite{al2019detecting,wahdan2022early,cechinel2021learning}. When identified, alerts prompt educators to investigate and provide necessary support.
In online learning environments like MOOCs, logistic regression and neural networks analyze clickstream data and forum participation to identify at-risk students \cite{choi2018learning,wahdan2022early,azcona2019detecting}. Early identification leads to targeted interventions, such as personalized messages and additional resources, improving retention rates. Similarly, in public schools, historical data on grades, attendance, and behavior predict dropout risks effectively, with targeted support programs significantly decreasing dropout rates \cite{baneres2019early,atif2020perceived}.
Interventions often include academic support programs like tutoring and study groups tailored to at-risk students. Personalized learning plans adapt the curriculum to individual learning paces, with regular check-ins to monitor progress. Behavioral and psychological support mechanisms, such as access to mental health professionals and motivational programs, enhance student engagement and address stress or anxiety \cite{trakunphutthirak2018detecting,hu2022using,yuan2021early}.

\subsubsection{Detecting Student Mental Health}
Mental health is a vital aspect of student well-being and academic success. Identifying students who may be experiencing psychological or emotional challenges involves analyzing various indicators beyond academic performance and behavioral data \cite{guo2022multimodal,baba2023prediction}. A crucial component of this task is considering psychosocial features, which offer direct insights into student mental and emotional state.
Academic performance can serve as an early warning system for mental health issues. Significant declines in grades often signal underlying psychological problems such as increased stress or anxiety \cite{sahlan2021prediction}. For example, a sharp fall in academic results might indicate heightened stress levels or other mental health concerns. Monitoring academic trends can help educators identify students who may need psychological support. 
Additionally, irregular attendance and decreased participation in class activities can be signs of emotional struggles. The integration of attendance data and academic metrics enables more accurate identification of students at risk \cite{ge2020research,shafiee2020prediction,yang2020using}. Psychosocial data, including stress levels, motivation, and emotional well-being, provides direct insights into student mental health and is typically gathered through surveys and self-reports \cite{guo2022multimodal,guo2022educational}. High levels of reported stress and decreasing academic performance often signal significant psychological distress. Integrating this psychosocial data with academic and behavioral data allows for a comprehensive understanding of mental health issues, facilitating more effective and timely interventions \cite{li2020identifying,lu2021impact}.

\begin{table*}[!htbp]
	\centering
	\caption{Summary of educational data types and models applied across various time series analysis scenarios in education.}\label{tabSummary}
	\renewcommand{\arraystretch}{1.2}
		\begin{threeparttable}
	\begin{tabular*}{\hsize}{@{}@{\extracolsep{\fill}}llcccccccc@{}}
		\toprule
		\multirow{3}{*}{} & \multirow{3}{*}{Type/Name} & \multicolumn{2}{c}{Forecasting} & \multicolumn{2}{c}{Classification} & \multicolumn{2}{c}{Clustering} & \multicolumn{2}{c}{Anomaly Detection}\\
		\cmidrule{3-4} 		\cmidrule{5-6}		\cmidrule{7-8} 		\cmidrule{9-10}
		&       & \makecell[c]{Grades and\\Success} & \makecell[c]{Enrollment} & 
		\makecell[c]{Learning\\Patterns} &
		\makecell[c]{Student\\Behavior} &
		\makecell[c]{Students} &
		\makecell[c]{Collaborative\\Learning} &
		\makecell[c]{At-Risk\\Learners} &
		\makecell[c]{Mental\\Health} \\
		\midrule
		\multirow{8}{*}{Data} &\makecell[l]{Academic\\Performance} & \usym{2714}      &    \usym{2714}   & \usym{2714}       &       &\usym{2714}        &       &\usym{2714}        &  \usym{2714} \\[8pt]
		&\makecell[l]{Learning\\Behavioral} &   \usym{2714}    &     & \usym{2714}      &    \usym{2714}     & \usym{2714}       &\usym{2714}        &    \usym{2714}    & \usym{2714}\\[8pt]
		&\makecell[l]{Socio-\\Demographic}     &\usym{2714}   & &       &  \usym{2714}      &       &       &  \usym{2714}      &  \\[8pt]
		& \makecell[l]{Psycho-\\social}&      \usym{2714}  & \usym{2714}       &  \usym{2714}      &       &       &  \usym{2714}      &       &\usym{2714} \\
		\midrule
		\multirow{11}{*}{Models} 
		&ARIMA	&$\CIRCLE$&	$\CIRCLE$&	$\LEFTcircle$&	$\LEFTcircle$&	$\CIRCLE$&$\LEFTcircle$&$\CIRCLE$&$\LEFTcircle$\\[3pt]
		&LSTM	&$\CIRCLE$&	$\LEFTcircle$&	$\CIRCLE$&	$\LEFTcircle$&	$\LEFTcircle$&$\CIRCLE$&$\CIRCLE$&$\LEFTcircle$\\[3pt]
		&SVM	&$\CIRCLE$&	$\LEFTcircle$&	$\CIRCLE$&	$\CIRCLE$&	$\CIRCLE$&$\LEFTcircle$&$\LEFTcircle$&$\CIRCLE$\\[3pt]
		&DT	&$\LEFTcircle$&	$\CIRCLE$&	$\CIRCLE$&	$\CIRCLE$&	$\LEFTcircle$&$\CIRCLE$&$\LEFTcircle$&$\LEFTcircle$\\[3pt]
		&RF	&$\CIRCLE$&	$\LEFTcircle$&	$\CIRCLE$&	$\CIRCLE$&	$\CIRCLE$&$\LEFTcircle$&$\CIRCLE$&$\CIRCLE$\\[3pt]
		&KNN	&$\LEFTcircle$&	$\LEFTcircle$&	$\CIRCLE$&	$\LEFTcircle$&	$\CIRCLE$&$\LEFTcircle$&$\LEFTcircle$&$\LEFTcircle$\\[3pt]
		&ANN &$\CIRCLE$&	$\CIRCLE$&	$\CIRCLE$&	$\CIRCLE$&	$\LEFTcircle$&$\LEFTcircle$&$\CIRCLE$&$\LEFTcircle$\\[3pt]
		&K-Means	&$\Circle$&	$\Circle$&	$\LEFTcircle$&	$\LEFTcircle$&	$\CIRCLE$&$\CIRCLE$&$\LEFTcircle$&$\CIRCLE$\\[3pt]
		&DBSCAN	&$\Circle$&	$\Circle$&	$\LEFTcircle$&	$\CIRCLE$&	$\CIRCLE$&$\CIRCLE$&$\LEFTcircle$&$\Circle$\\
		\bottomrule
	\end{tabular*}%
		\begin{tablenotes}
	\item $\Circle$ represents “less frequently used”, $\LEFTcircle$ represents “normally used”, and $\CIRCLE$ represents “frequently used”.
\end{tablenotes}
\end{threeparttable}
\end{table*}

Various strategies are increasingly employed to detect mental health issues at an early stage. Regression methods are commonly used to model and understand the temporal dynamics of indicators related to student well-being \cite{bergmann2020uninformed}. For instance, linear regression models analyze how deviations from expected patterns in grades or participation might correlate with stress or depression indicators \cite{ren2021deep}. Researchers use these models to assess how changes in predictors, such as academic metrics and attendance rates, relate to outcomes indicating mental health issues \cite{lu2021anomaly,laxhammar2013online}.
Multivariate regression extends this approach by considering multiple predictors simultaneously, integrating various data types such as academic performance, social interactions, and behavioral observations \cite{russell2020elements,vaidya2024anomaly}. This method reveals that a combination of these indicators is a more robust predictor than any single factor alone. In educational settings, regression techniques help identify patterns and predict potential mental health concerns by analyzing changes in student performance and behavior over time \cite{norouzi2023context}.
Additionally, forecasting models explore physiological data to predict mental health issues. Predictive modeling techniques, including traditional statistical methods and deep learning, integrate physiological indicators to forecast potential mental health deteriorations \cite{li2023retracted,pabalkar2023anomaly,li2023simulation}. Recent advancements focus on multimodal approaches that combine academic records, behavioral data, and physiological indicators to create a comprehensive view of student mental health \cite{cechinel2024lanse,guo2022educational}. By integrating these diverse data sources, studies have improved detection capabilities and provided a more nuanced understanding of mental health trends.

\subsection{Summary}
Table \ref{tabSummary} summaries the various types of educational data and the models across different time series analysis tasks in education. Regarding the data, academic performance data and learning behavioral data are the most heavily utilized, covering nearly all applied scenarios. Academic performance data is particularly crucial for forecasting tasks, such as predicting grades and enrollment trends, which are essential for planning educational interventions and identifying at-risk students through anomaly detection. Learning behavioral data is instrumental in classification and clustering tasks, helping to categorize students based on their engagement levels, learning styles, and behavioral traits, which can inform personalized teaching strategies. Socio-demographic data plays a specialized role, primarily in forecasting and anomaly detection, where it helps to identify potential challenges faced by at-risk learners. Psychosocial metrics, meanwhile, are extensively used in forecasting and classification, uncovering details of  how psychological and social factors influence student outcomes. These metrics are also significant in clustering for collaborative learning and detecting mental health issues, emphasizing their importance in understanding the broader context of student performance.

In terms of models, ARIMA and LSTM are the most frequently used for forecasting tasks, particularly in predicting grades and enrollment, with LSTM being especially effective due to its ability to capture long-term dependencies in sequential data. SVM and RF are versatile models, frequently applied across classification, clustering, and anomaly detection tasks. SVM is robust in handling high-dimensional data, making it suitable for classifying learning patterns and student behavior, while ensemble nature of RF enhances its accuracy across these tasks. K-Means and DBSCAN are prominent in clustering applications, with K-Means being preferred for grouping students and collaborative learning due to its simplicity, while DBSCAN excels in handling noise and outliers. ANN is widely used for both forecasting and classification, excelling in complex pattern recognition tasks, while DT is valued for its interpretability, particularly in classifying student behavior. These models are strategically employed to address the diverse analytical needs in education, offering tailored solutions for forecasting, classification, clustering, and anomaly detection, depending on the specific characteristics of the educational task at hand.

\section{Educational Scenarios and Applications}\label{secApplication}
In Section \ref{secMethod}, we provide a detailed review of various time series models in education from a methodological perspective. This section then shifts focus to their applications in educational scenarios, and demonstrate how different time series techniques are applied in educational tasks.
By examining concrete scenarios, we aim to present a thorough overview of the adaptability and impact of time series analysis in education.

\subsection{Academic Performance Prediction}
Predicting academic performance is a pivotal task in educational settings, allowing institutions to foresee student outcomes and intervene proactively to support students in need \cite{zhang2021educational,yaugci2022educational,waheed2020predicting,xu2017progressive}. Time series methods analyze sequential data points, such as grades, attendance records, and participation rates, collected over time, enabling educators to forecast future academic outcomes based on historical trends. One of the most robust forecasting models is the ARIMA model, widely used due to its capacity to handle various types of time series data effectively by considering past performance and trends \cite{qin2019impact,batool2023educational,albreiki2021systematic}. Additionally, machine learning and deep learning models are popular for their ability to learn long-term dependencies and provide smooth and accurate forecasts, making them suitable for complex academic data influenced by numerous factors over time \cite{chen2023comparative,sekeroglu2019student,kim2018gritnet}. Classification models also play a crucial role by measuring similarities between sequences that may vary over time, helping to identify students with similar learning patterns \cite{dobashi2022learning,haiyang2018time}. They classify student performance states (e.g., high, medium, low) based on observed sequences of academic achievements and transitions over time, providing a deeper understanding of performance dynamics.

Combining these methods offers a comprehensive approach to predicting academic performance. Forecasting models like ARIMA or LSTM can predict future grades, which are then classified into performance categories using classification strategies \cite{pandey2016towards}. Clustering techniques like k-means can group students based on these predictions, enabling targeted interventions for each cluster \cite{perera2008clustering}. When comparing these methods, forecasting models provide precise numerical predictions of future performance, while classification models categorize students into distinct performance levels, making it easier to design targeted interventions \cite{deeva2022predicting}. 
Time series analysis offers multiple approaches to academic performance prediction, and integrating these methods enhances the robustness of predictions. The main advantage lies in its ability to handle sequential data, capturing trends and patterns that static methods might miss. By combining these methods, educational institutions can not only predict future performance but also classify and group students effectively, ensuring tailored support and timely interventions \cite{al2019detecting,choi2018learning}. Overall, the application of time series methods in academic performance prediction demonstrates significant potential in transforming educational practices, highlighting the importance of continuous innovation and the integration of advanced analytical techniques in education.

\subsection{Analyzing Learning Behaviors}
Understanding and analyzing student behavior is essential for fostering a positive learning environment and enhancing educational outcomes. Classification models are at the forefront of analyzing learning behaviors due to their ability to categorize complex behavioral patterns effectively. Decision tree classifiers, for instance, categorize student behaviors such as participation, submission punctuality, and online activity levels \cite{dutsinma2020vark,sheeba2018prediction}. By identifying and classifying these behaviors, educators can tailor interventions to address specific issues, such as providing additional resources to students who consistently submit assignments late or offering motivational support to those who show declining participation. Dynamic time warping is another useful classification technique, measuring similarities between sequences of behavioral data to identify typical or atypical learning patterns \cite{shen2017clustering,he2023clustering}. These classifications help in creating targeted support strategies that cater to individual student needs.

Forecasting models complement classification methods by predicting future behavioral trends and engagement levels. Exponential smoothing techniques, for example, can forecast student engagement by analyzing past interaction data from learning management systems \cite{shingari2017review}. LSTM can be used to forecast engagement trends to provide accurate predictions of future behavior, which are then classified into specific behavior categories using decision tree classifiers \cite{bai2021educational}. 
In addition, clustering models further enhance the analysis of learning behaviors by grouping students based on their classified behavioral patterns. Hierarchical clustering can identify students with similar engagement and participation trends, allowing educators to design group-specific interventions \cite{bueno2023hierarchical,akhanli2024hierarchical}. For example, students displaying similar patterns of disengagement might benefit from collaborative learning activities tailored to re-engage them. 
Forecasting models predict future engagement levels, while classification models categorize behaviors to tailor support strategies. Clustering techniques group students with similar behavioral patterns, enabling targeted interventions for each cluster. Anomaly detection models ensure that individual deviations are promptly identified and addressed, maintaining continuous support for all students.

\subsection{Learner Profiling and Modeling}
Learner profiling and modeling include creating detailed profiles of students based on their learning behaviors, preferences, and performance. This process helps in understanding individual learning needs and tailoring educational strategies accordingly. Time series analysis in this task provides the temporal dynamics of learner behavior, enabling the creation of robust and dynamic profiles. Among the various time series methods, clustering models, particularly k-means clustering, are most widely used for learner profiling and modeling \cite{perera2008clustering,pasina2019clustering}. Specifically, k-means is extensively used to group learners based on their behavioral patterns captured over time \cite{chi2021research,moubayed2020student,kuswandi2018k}. This method involves partitioning students into clusters based on the similarities in their time series data, such as grades, participation, engagement metrics, and other academic activities. By applying k-means clustering to these data points, educators can identify distinct groups of learners with similar learning trajectories and needs. For instance, the method can be applied to time series data of students grades across different subjects over a semester \cite{trivedi2020clustering}. This clustering can reveal groups of students who consistently perform well, those who show gradual improvement, and those who may be struggling.

The clustering models are also combined with other time series methods to enhance learner profiling. For instance, after clustering students based on their engagement patterns, anomaly detection models can be applied within each cluster to identify students whose behavior deviates significantly from the cluster norm \cite{guo2022educational,vaidya2024anomaly}. This approach ensures that students who might be struggling within a generally engaged cluster are not overlooked, allowing for timely and targeted interventions. Besides, forecasting models can be used to predict future engagement levels and academic outcomes within each cluster identified by k-means \cite{trivedi2020clustering,cahapin2023clustering}. For instance, LSTM can forecast the future performance of students in a cluster of high achievers.
After clustering students, classification models are applied to classify individual behaviors within each cluster. For example, within a cluster of sporadically engaged students, it can classify specific engagement patterns, such as students who log in irregularly but participate intensely when they do \cite{xu2019automatic,norouzi2023context,laxhammar2013online}.
Overall, by integrating these methods, a comprehensive approach to learner profiling and modeling is achieved. For instance, forecasting models provide insights into future trends, allowing for proactive educational planning, while anomaly detection models maintain continuous monitoring, ensuring that any significant deviations are quickly identified and addressed.

\subsection{Summary}
In this section, we have systematically organized and classified various educational applications and scenarios of time series analysis based on their purposes and characteristics. Building upon the specific applications discussed in Section \ref{secMethod}, we provide a comprehensive overview that encompasses the majority of educational scenarios. Previous surveys have classified educational tasks and applications from various perspectives according to their different objectives. However, the boundaries between these applications and the methods used are often blurred. For instance, while academic performance prediction is typically associated with forecasting, it can also serve classification tasks, such as categorizing student performance based on predicted outcomes. Similarly, models and methods commonly employed in forecasting, such as ARIMA and LSTM, are frequently used in classification tasks, highlighting the interconnected nature of these applications.

\begin{table}[!htbp]
	\centering
	\caption{The summary of integration of different time series techniques applied in mainstream educational scenarios and applications .}\label{tabApplication}
	\renewcommand{\arraystretch}{2}
\begin{threeparttable}
	\begin{tabular*}{\hsize}{@{}@{\extracolsep{\fill}}lcccc@{}}
		\toprule
		Methods& \makecell[c]{Forecasting} & \makecell[c]{Classifying} & \makecell[c]{Clustering}&\makecell[c]{Detecting} \\ 
		\midrule
        \makecell[l]{Academic\\Performance\\Prediction}&\CIRCLE&\CIRCLE&\LEFTcircle&\Circle\\[12pt]
        \makecell[l]{Analyzing\\Learning\\Behaviors}&\CIRCLE&\CIRCLE&\CIRCLE&\LEFTcircle\\[12pt]
        \makecell[l]{Learner\\Profiling and\\Modeling}&\LEFTcircle&\CIRCLE&\CIRCLE&\CIRCLE\\
		\bottomrule
	\end{tabular*}
\begin{tablenotes}
	\item $\Circle$ represents “less frequently used”, $\LEFTcircle$ represents “normally used”, and $\CIRCLE$ represents “frequently used”.
\end{tablenotes}
\end{threeparttable}
\end{table}

In response to this, we integrate these applications from a time series perspective and explore how multiple time series techniques are combined and applied within educational settings, as summarized in Table \ref{tabApplication}. The integration of these methods across various educational scenarios underscores their significant impact on improving educational outcomes. When used in combination, these techniques provide a more thorough understanding of student performance and engagement, facilitating the development of more responsive and adaptive educational environments. Such environments are better equipped to meet the diverse needs of students, ultimately leading to improved academic success and well-being.
As the field of educational data analysis continues to advance, the ongoing development and application of these models will be critical in addressing the increasingly complex challenges in education. This integrated approach not only enhances our understanding of how time series methods are applied in different educational scenarios but also emphasizes the synergistic use of these methods to improve educational outcomes.

\section{Future Directions}\label{secFuture}
\subsection{Personalized Learning Analytics}
Leveraging time series data to create personalized learning pathways can significantly improve student outcomes by providing a detailed analysis of individual learning patterns and progress over time. This comprehensive data allows educators to customize instructional strategies to better meet the unique needs of each student, offering targeted support and resources precisely when required. For instance, time series data can highlight specific moments when a student begins to struggle with certain concepts, enabling educators to intervene promptly and effectively \cite{shoaib2024ai,hong2024early}. This proactive approach ensures that students do not fall behind and receive the help they need exactly when they need it.
Furthermore, the development of adaptive learning systems that dynamically adjust the learning content and pace based on real-time data analysis will be a crucial area of research \cite{choi2018learning,shoaib2024ai,li2024bringing}. These advanced systems have the capability to monitor student interactions with educational material continuously, identify areas where they excel or face challenges, and adapt the curriculum accordingly. By providing personalized feedback and recommendations, adaptive learning systems make the learning process more effective and engaging for each student \cite{imhof2020implementation}. For example, if a student consistently struggles with a particular topic, the system can offer additional resources or modify the instructional approach to better suit the student’s learning style.

Integrating adaptive learning systems with time series analysis fosters a continuous feedback loop, wherein data-driven insights lead to immediate instructional adjustments \cite{vsaric2023student}. This synergy not only enhances the overall learning experience but also maximizes the potential for academic achievement. Such systems ensure that students remain appropriately challenged and supported, promoting a more efficient and tailored educational experience. Future research should focus on refining these adaptive systems to better predict student needs and optimize learning outcomes \cite{chango2021improving}. By advancing these technologies, educators can ensure that every student benefits from a truly personalized education, paving the way for more individualized and effective learning environments. This holistic approach to personalized learning pathways holds great promise for transforming education, making it more responsive to the needs of each learner.

\subsection{Multimodal Data Fusion}
The integration of multimodal data in educational settings offers a transformative potential to enhance the understanding of student learning and performance. By combining different types of data, such as time series data, text data, video data, and sensor data, educators and researchers can create comprehensive student profiles that provide an overall assessment of student behavior and academic performance \cite{chango2022review,henderson2020improving,mu2020multimodal}. Time series data, for example, includes sequential information like attendance records and grades over time, which can be complemented by textual analysis from essays and discussion forums. Video data, capturing classroom interactions and lectures, alongside sensor data from wearables, add rich contextual layers to these profiles \cite{prieto2018multimodal,chango2021multi}.

The primary advantage of multimodal data integration lies in its ability to improve the predictive power of educational models. By leveraging multiple data sources, predictive models can achieve higher accuracy, leading to more precise and actionable insights. This approach not only enhances the ability to forecast student outcomes but also provides a more nuanced understanding of the learning process. For instance, combining behavioral data with academic performance metrics can identify early signs of disengagement or potential dropout risks, allowing for timely interventions \cite{guo2022multimodal,li2020identifying,jin2024survey}. However, the integration of multimodal data is not without challenges. One significant issue is data heterogeneity, as different data types often come in various formats and structures \cite{mu2020multimodal,zhao2023construction}. Developing standardized frameworks and protocols for data integration is crucial to address this challenge. Another concern is the volume and complexity of data, which requires advanced data storage and processing technologies like big data analytics \cite{shankar2019architecture}. Temporal alignment of data streams collected at different times is also essential, necessitating sophisticated time alignment algorithms to ensure data consistency.

\subsection{LLMs in Educational Time Series}
The advent of LLMs presents a novel opportunity to revolutionize the use of time series analysis in the educational sector~\cite{xu2024foundation,li2023adapting,zhang2024self,jin2023large}. These models, known for their ability to understand and generate human-like text, can be harnessed in conjunction with time series data to create more sophisticated and dynamic educational tools. For instance, LLMs can be integrated with time series data to enhance predictive analytics, allowing for more nuanced predictions about student outcomes \cite{chang2024survey,dan2023educhat}. By analyzing patterns in student performance, behavior, and engagement over time, LLMs can identify which students might need additional support, thus enabling proactive interventions. This integration can also significantly enhance personalized learning experiences by tailoring educational content to fit individual learning styles and preferences based on engagement levels and performance trends.

Furthermore, LLMs can play a crucial role in developing advanced early warning systems and intelligent tutoring systems within the context of educational time series data \cite{mousavinasab2021intelligent,jin2024position}. These systems can utilize LLMs to interpret time series data and provide contextually relevant feedback, explanations, and hints, maintaining student engagement and improving understanding. Additionally, the fusion of large-scale data and multimodal data can offer a comprehensive view of student learning dynamics, providing deeper insights for curriculum development and real-time feedback. However, the integration of LLMs and time series data should be approached with careful consideration of ethical implications, ensuring data privacy, security, and fairness to maintain trust and promote equitable treatment across diverse student populations \cite{gan2023large,yan2024practical}. The potential of LLMs to transform educational practices and outcomes through sophisticated analysis and personalized support is immense, paving the way for more adaptive and effective learning environments.

\subsection{Scalability and Generalization}
As educational datasets increase in size and complexity, the scalability of time series analysis methods becomes crucial. Processing large volumes of data while maintaining performance is a significant challenge that educators and data scientists need to address \cite{ang2020big,zhang2024hyperscale}. One effective solution involves leveraging distributed computing and cloud-based platforms, which offers the computational power necessary to handle extensive datasets, ensuring that time series analysis remains efficient even as data scales up \cite{mitra2020mobile,govea2023optimization}. Additionally, big data frameworks can also facilitate the processing and analysis of massive educational datasets. These frameworks distribute data processing tasks across multiple nodes, significantly speeding up the analysis and making it feasible to work with very large datasets \cite{julia2021educational}. 

Beyond scalability, the ability of time series analysis methods to generalize across different educational contexts and diverse student populations is equally essential. Effective time series models should adapt to varying curriculum structures, cultural differences, and educational systems. Techniques such as transfer learning and domain adaptation are invaluable in this regard \cite{hunt2017transfer,li2023holistic}. Transfer learning, for instance, leverages a pre-trained model on a similar task, fine-tuning it with data from the new context \cite{zhuang2020comprehensive}. This approach reduces the amount of data and time required for training while maintaining accuracy. On the other hand, domain adaptation strategies adjust models to new environments without extensive retraining, ensuring adaptability and efficiency \cite{fu2020learning,barbosa2024adaptive}.
These approaches are particularly useful when applying a model trained on urban schools to rural schools, where demographic and resource differences might otherwise necessitate a complete retraining of the model.

\subsection{Cross-Disciplinary Collaboration}
Cross-disciplinary collaboration is essential for advancing the use of time series analysis in education, as it unites expertise from diverse fields such as education, data science, psychology, and sociology \cite{romero2010educational,hussain2011affect}. By leveraging insights from these various disciplines, we can develop more comprehensive and effective time series models that address the multifaceted nature of educational data. For instance, educators can provide contextual knowledge about pedagogical practices and learning environments, which can inform the design of more relevant and accurate time series models \cite{xie2024co,romero2017educational}. Data scientists can contribute advanced analytical techniques and algorithms to process and interpret complex educational datasets, while psychologists and sociologists can help explain the behavioral and social patterns that emerge from time series analyses, leading to more inclusive interpretations and applications \cite{hussain2018educational,tran2020toward,dutsinma2020vark}.

Creating collaborative platforms such as interdisciplinary research centers, virtual symposiums, and collaborative software tools can facilitate seamless communication and knowledge sharing among researchers from different fields \cite{olsen2020temporal}. These platforms can support the integration of diverse datasets, enabling the combination of time series data with other types of educational data, such as text, video, and demographic information. This multimodal approach can uncover deeper insights into student behavior and learning processes, providing a more comprehensive understanding of educational dynamics \cite{chango2021multi}.
Besides, collaborative tools and technologies, such as shared databases, data visualization tools, and collaborative software platforms, can enhance the effectiveness of cross-disciplinary work on time series analysis \cite{li2023retracted,cechinel2024lanse}. These tools can support the integration and analysis of time series data from multiple sources, enabling real-time data sharing and collaborative problem-solving. For example, interactive data visualization tools can help researchers from different disciplines explore time series data together, uncovering patterns and insights that might not be apparent through traditional analysis methods \cite{nguyen2021interactive,romero2020educational}.

\section{Conclusion}\label{secConclusion}
This paper has provided an in-depth exploration of the role of time series analysis in education, highlighting its wide-ranging methods, applications, and future research directions. Time series analysis serves as a powerful tool for uncovering temporal patterns in educational data, empowering educators and researchers to make informed, data-driven decisions that can significantly enhance learning outcomes. As the field continues to evolve, driven by the proliferation of digital learning platforms and the growing availability of big data, the integration of multimodal data and the development of more robust, scalable models will be essential. Cross-disciplinary collaboration, drawing from education and data science, will further refine these models that more accurately capturing the complexities of learning environments.

The future of time series analysis in education is promising, with potential breakthroughs in personalized learning and adaptive educational systems. By incorporating diverse data sources and advanced analytical techniques, educators can develop more tailored interventions that address individual student needs more effectively. As these methodologies become increasingly sophisticated and widely adopted, their application will not only improve educational outcomes but also contribute to the creation of more equitable and responsive learning environments across various educational settings, fostering inclusivity and supporting lifelong learning.

%


\bibliographystyle{IEEEtran}
\bibliography{refs}

\end{document}